\documentclass[sigconf,nonacm]{acmart}

\AtBeginDocument{%
  }

\setcopyright{none}
\begin{document}

\title{MultiCompose: Multi-Concept Personalized Composition with Per-Subject Attribute Binding}

\author{Ruirui Zhang}
\orcid{0009-0009-9877-1818}
\authornote{Both authors contributed equally to this research.}
\affiliation{%
  \institution{Nanjing University of Aeronautics and Astronautics}
  \department{College of Computer Science and Technology}
  \city{Nanjing}
  \state{Jiangsu}
  \country{China}}
\email{qsswhh@nuaa.edu.cn}

\author{Zhengkai Zhao}
\orcid{0009-0003-5686-3868}
\authornotemark[1]
\affiliation{%
  \institution{Nanjing University of Aeronautics and Astronautics}
  \department{College of Artificial Intelligence}
  \city{Nanjing}
  \state{Jiangsu}
  \country{China}}
\email{zhengkai.zhao@nuaa.edu.cn}

\author{Pan Gao}
\orcid{0000-0002-4492-5430}
\correspondingauthor
\affiliation{%
  \institution{Nanjing University of Aeronautics and Astronautics}
  \department{College of Artificial Intelligence}
  \city{Nanjing}
  \state{Jiangsu}
  \country{China}}
\email{pan.gao@nuaa.edu.cn}


\begin{abstract}
Text-to-image diffusion models enable personalization of specific visual concepts from a small number of reference images.
However, generating a single image that contains multiple personalized
subjects, each bound to user-specified attributes
such as clothing, accessories, and held objects, remains largely unaddressed.
Without explicit spatial constraints, concurrently activated concept
checkpoints produce overlapping cross-attention responses, causing
per-subject identity degradation and attribute misalignment.
Moreover, no established benchmark jointly evaluates these two failure modes
in the personalized multi-subject setting.
We present MultiCompose, a composition framework that decouples
per-concept personalization from multi-subject inference.
A semantic preservation regularization maintains attribute binding
capacity during fine-tuning, while a two-phase inference procedure
automatically establishes subject layout and composes per-concept
predictions through spatially exclusive masks.
We further introduce MSP-Bench, a benchmark that jointly evaluates
identity fidelity (ID), attribute binding accuracy (BIND), and attribute misalignment (MIS) through
a dual-pathway protocol.
Experiments show that MultiCompose outperforms existing methods
on both conventional metrics and MSP-Bench, confirming the benchmark's
ability to reveal failure modes that conventional metrics overlook.
Code is available at \url{https://github.com/I2-Multimedia-Lab/MultiCompose}.
\end{abstract}


\begin{teaserfigure}
  \centering
  \includegraphics[width=\textwidth]{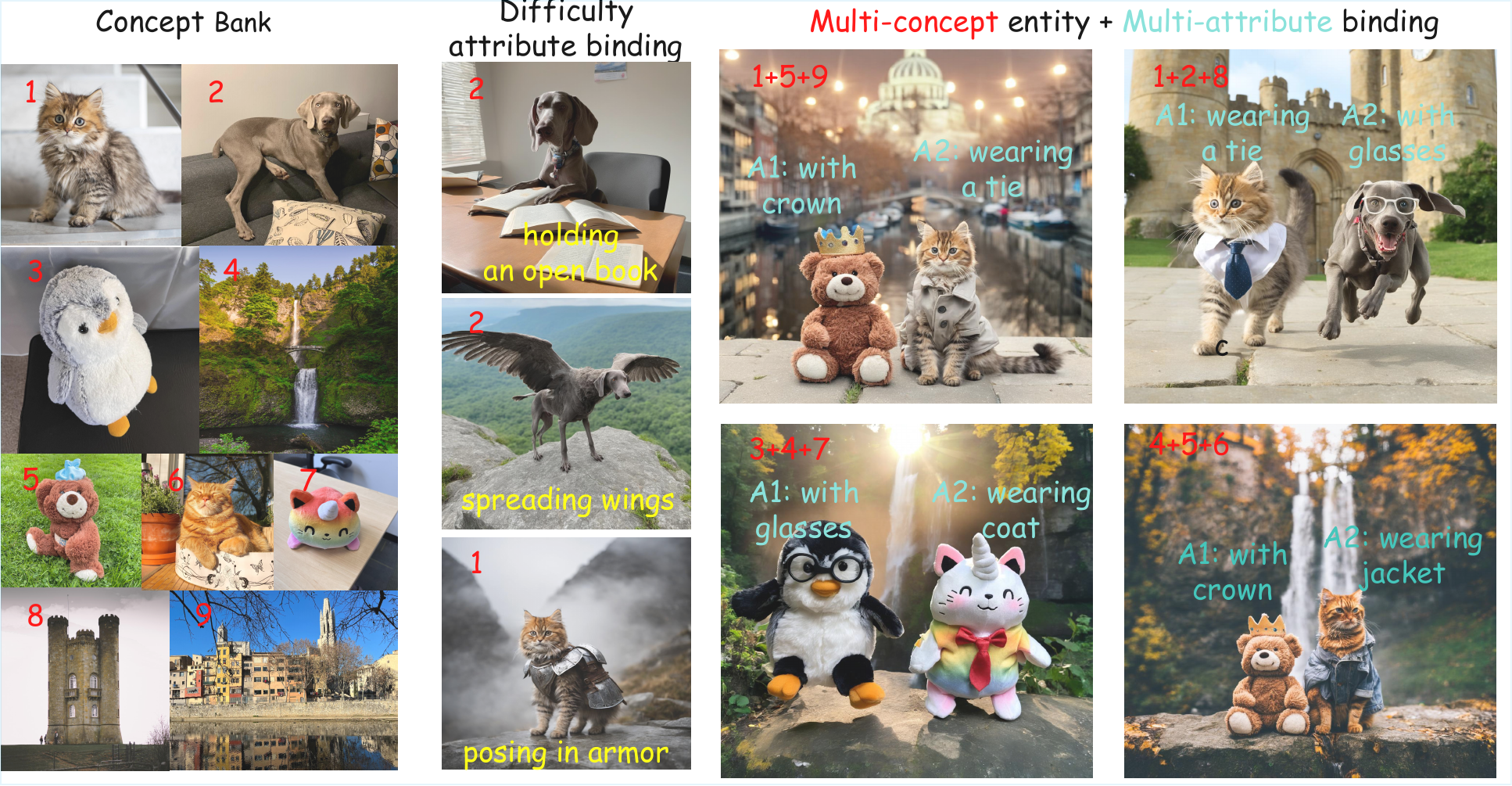}
  \caption{\textbf{Multi-subject personalized generation with attribute binding.}
  MultiCompose composes independently fine-tuned concepts into a single image
  while binding user-specified attributes exclusively to their designated subjects.
  Left: single-concept generation with difficult attribute binding
  (e.g., specific accessories or held objects).
  Right: multi-concept composition where each personalized subject
  retains its learned identity and receives its specified attributes
  without cross-subject leakage.}
  \Description{Examples of single-subject and multi-subject personalized
  generation. The multi-subject examples preserve each subject's identity
  and assign attributes to the designated subject without cross-subject
  leakage.}
  \label{fig:teaser}
\end{teaserfigure}

\maketitle

\section{Introduction}

Text-to-image (T2I) diffusion models~\cite{rombach2022ldm, podell2023sdxl,saharia2022photorealistic}
have enabled high-quality image synthesis from natural language descriptions.
A significant line of work extends this capability to \emph{personalized}
generation, where a model is adapted to reproduce a specific visual concept
from a small reference set~\cite{ruiz2023dreambooth, gal2022textual,
hu2022lora}.
These methods successfully generate novel scenes containing a single
personalized subject and have found broad application in story visualization,
game asset design, advertising, and personalized portraiture.

Despite this progress, a practically important scenario remains largely
unaddressed: generating a single image that simultaneously contains
\textbf{multiple personalized subjects}, each bound to \textbf{user-specified
attributes} such as clothing, accessories, and held objects.
This task imposes two requirements that existing methods do not jointly
satisfy: each subject must preserve its fine-grained identity as learned
from reference images, and each attribute must be correctly and exclusively
bound to its designated subject~\cite{feng2023trainingfree,tang2023daam}.
Directly combining independently fine-tuned checkpoints fails to meet
these requirements: without explicit constraints on cross-attention
responses, concurrent activation of multiple concept checkpoints
causes \textbf{per-subject identity degradation}, and
attention diffusion across text tokens produces \textbf{attribute
misalignment}, where visual properties are incorrectly transferred
between subjects.

Existing approaches fall into two categories: methods that require
joint training across all concepts to achieve multi-subject
composition~\cite{gu2024mixofshow}, and methods that provide spatial
layout control but operate only on non-personalized
content~\cite{bar2023multidiffusion, avrahami2023spatext,zhang2023adding,li2023gligen}.
Neither category supports composing independently personalized subjects
with per-subject attribute binding.
Furthermore, existing benchmarks evaluate either single-concept
identity preservation or attribute binding in non-personalized settings,
but none jointly evaluates these two failure modes in the personalized
multi-subject setting, leaving this task without a suitable
evaluation protocol.

We present \textbf{MultiCompose}, a composition framework that separates
per-concept personalization from inference-time multi-subject composition.
Per-concept weight offsets learned during independent fine-tuning
determine subject identity, while attribute binding is handled entirely
at inference time through spatial routing of cross-attention and
mask-guided noise prediction composition,
eliminating the need for joint training across concepts.
Specifically, MultiCompose consists of three components.
(1) During fine-tuning, a \textbf{regularization term} constrains each
modifier token embedding to remain close to its category word,
preserving attribute-to-subject binding capacity for personalized concepts.
(2) At inference, a \textbf{pre-fusion} phase uses a global prompt with
token-level cross-subject attention isolation to establish the spatial
layout of all subjects while preventing cross-subject attribute leakage,
without requiring any user-provided spatial prior such as bounding boxes.
A subsequent \textbf{fusion} phase activates per-concept weight sets under
their respective single-subject prompts and composes the resulting noise predictions
through mutually exclusive soft masks, so that each subject's identity
and attributes are spatially confined to its designated region.
(3) We further introduce \textbf{MSP-Bench}, a benchmark extending
T2I-CompBench++~\cite{huang2023t2icompbench} to the personalized
multi-subject setting, providing joint automatic assessment of identity
fidelity (ID), attribute binding accuracy (BIND), and attribute misalignment (MIS).
Fig.~\ref{fig:teaser} shows representative results.
The contributions of this work are as follows:
\begin{itemize}
\item \textbf{Personalized fine-tuning:} We identify that
modifier token embeddings diverge from their category words during
fine-tuning, and propose an embedding regularization term to preserve
semantic proximity, enabling correct attribute binding to
personalized subjects.
\item \textbf{Inference-time composition:} We propose \textbf{MultiCompose},
a two-phase inference framework that composes multiple independently
personalized subjects via token-level attention isolation for layout
initialization and mask-guided noise prediction routing, without
joint training across concepts or user-provided spatial priors.
\item \textbf{Evaluation benchmark:} We introduce \textbf{MSP-Bench}, a benchmark
that jointly assesses identity fidelity (ID), attribute binding accuracy (BIND),
and attribute misalignment (MIS), providing a suitable evaluation protocol for
multi-subject personalized generation.
\end{itemize}

\section{Related Work}

\subsection{Text-to-Image Diffusion Models}
Denoising diffusion probabilistic models~\cite{ho2020ddpm} established
the theoretical foundation for iterative noise-to-image generation.
Latent diffusion models~\cite{rombach2022ldm} built on this framework
by moving the denoising process into a compressed latent space,
reducing computational cost while maintaining high visual fidelity.
Subsequent studies reveal that cross-attention layers
serve as the primary interface between text and visual features:
Prompt-to-Prompt~\cite{hertz2022prompt} shows that the spatial
layout of generated content is directly governed by per-token
cross-attention maps, enabling precise text-driven control over image
synthesis~\cite{chefer2023attend,li2023divide,rassin2024linguistic}.
Building on this understanding, SDXL~\cite{podell2023sdxl} further
advances generation quality through architectural scaling and a
dual-encoder text conditioning scheme that strengthens semantic
alignment between prompts and generated images~\cite{agarwal2023star,feng2023trainingfree,ge2023expressive}.
MultiCompose adopts SDXL as the backbone and routes per-subject and
per-attribute attention at inference time without modifying model weights.

\subsection{Personalized Generation and Multi-Subject Composition}
Personalized generation adapts a pretrained T2I model to reproduce a
specific visual concept from a small reference set.
Textual Inversion~\cite{gal2022textual} optimizes a new token embedding while
freezing all model weights.
DreamBooth~\cite{ruiz2023dreambooth} improves fidelity by
fine-tuning the full network with a prior preservation loss, at the cost
of per-concept weight updates.
CustomDiffusion~\cite{kumari2023customdiffusion} reduces this
overhead by updating only the cross-attention projections,
enabling more efficient adaptation across multiple concepts~\cite{han2023svdiff,Tewel2023KeyLockedRO}.
Yet none of these methods addresses the identity degradation and
attribute misalignment that arise under concurrent multi-concept activation~\cite{Xie_2023_ICCV,wang2024compositional}.
Composing multiple personalized concepts into a single image requires
mechanisms beyond independent fine-tuning.
Mix-of-Show~\cite{gu2024mixofshow} addresses this by merging concepts via
gradient fusion at training time, though this requires all concepts to be
known in advance.
TweedieMix~\cite{kwon2024tweediemix} and ConceptWeaver~\cite{kwon2024concept}
blend per-concept noise predictions in the denoised image space.
However, without explicit attention routing, these methods remain susceptible to semantic
leakage between subjects~\cite{li2024mulan,yang2024mastering,hu2024ella,jiang2024comat}.
In contrast, MultiCompose requires no joint training and addresses both
spatial and semantic interference through automatically detected soft masks and
token-level cross-subject attention isolation within a two-phase
inference procedure, without user-provided spatial priors.

\subsection{Evaluation Benchmarks for Compositional Generation}
Two lines of benchmarks are relevant to this work, yet neither covers
the personalized multi-subject setting.
T2I-CompBench++~\cite{huang2023t2icompbench} evaluates compositional
text-to-image generation across attribute binding, spatial relationships,
and non-spatial relations, but operates entirely outside the personalized
setting.
DreamBench++~\cite{peng2024dreambench} targets single-concept
personalization, assessing concept fidelity and prompt alignment,
but does not consider multi-subject composition or attribute binding.
Neither benchmark jointly evaluates identity fidelity and
compositional attribute binding in the personalized multi-subject setting,
leaving this task without an adequate evaluation protocol.
We address this gap by introducing a benchmark that
extends T2I-CompBench++ with identity fidelity metrics tailored to
the multi-subject personalized setting.

\section{Method}

\subsection{Problem Formulation}
\label{sec:formulation}

\begin{figure}[t]
  \centering
  \includegraphics[width=0.9\linewidth]{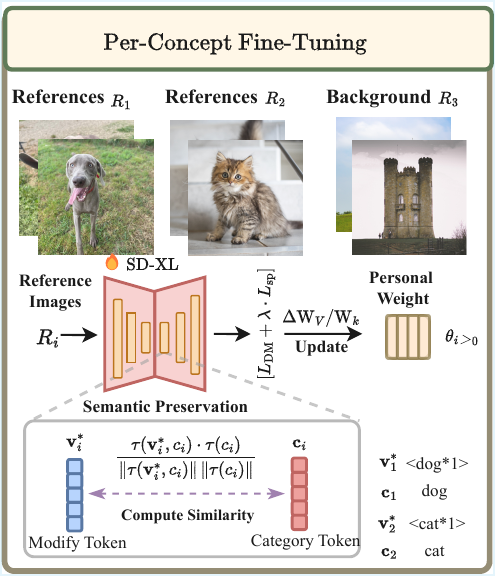}
  \caption{\textbf{Per-concept fine-tuning with semantic preservation.}
  The regularization term $L_{\mathrm{sp}}$ (Eq.~\ref{eq:spl}) constrains
  the modifier phrase embedding to remain close to the category word
  embedding, preserving attribute binding capacity.}
  \Description{A diagram of independent per-concept fine-tuning with a
  semantic preservation regularizer that keeps each modifier phrase close
  to its category embedding.}
  \label{fig:finetune}
\end{figure}

Let $\theta_0$ denote the frozen parameters of a pretrained text-to-image diffusion model~\cite{rombach2022ldm,podell2023sdxl}.
Given $N$ independently personalized concepts, concept $i \in \{1,\dots,N\}$ is parameterized
by per-concept updates to the key and value projection matrices
$\{\Delta W^K_{i,l}, \Delta W^V_{i,l}\}$ across all cross-attention layers of the denoising
U-Net, together with a learnable modifier token embedding $\mathbf{v}^*_i$ whose
corresponding category word is $c_i$~\cite{kumari2023customdiffusion}
(Fig.~\ref{fig:finetune}).
Each concept is fine-tuned independently with no gradient exchange between concepts~\cite{han2023svdiff,Tewel2023KeyLockedRO}.
We write $\theta_i$ to denote $\theta_0$ with concept $i$'s weight updates applied.
At inference the user provides:
(i)~per-subject attribute sets $\{A_i\}_{i=1}^N$, where each $A_i$ lists attribute
descriptions to be bound exclusively to subject $i$;
and (ii)~per-concept reference image sets $\{R_i\}_{i=1}^N$ used for evaluation.
No spatial layout specification (e.g., bounding boxes) is required from the user;
subject positions emerge from the pre-fusion phase and are detected automatically
(Sec.~\ref{sec:inference}).
The generation target is a single image in which each subject faithfully reflects the
visual identity encoded in $\theta_i$ and $\mathbf{v}^*_i$, and each attribute in $A_i$
is rendered correctly and exclusively on subject~$i$.

MultiCompose addresses this through two components:
(1)~a per-concept fine-tuning procedure with semantic preservation regularization, combined
with a two-phase inference procedure for spatial disentanglement;
and (2)~\textbf{MSP-Bench}, a benchmark for joint evaluation of subject identity fidelity,
cross-subject identity isolation, and compositional attribute binding.

\subsection{MultiCompose: Regularization and Inference-time Composition}
\label{sec:MultiCompose}

\begin{figure*}[t]
  \centering
  \includegraphics[width=0.95\linewidth]{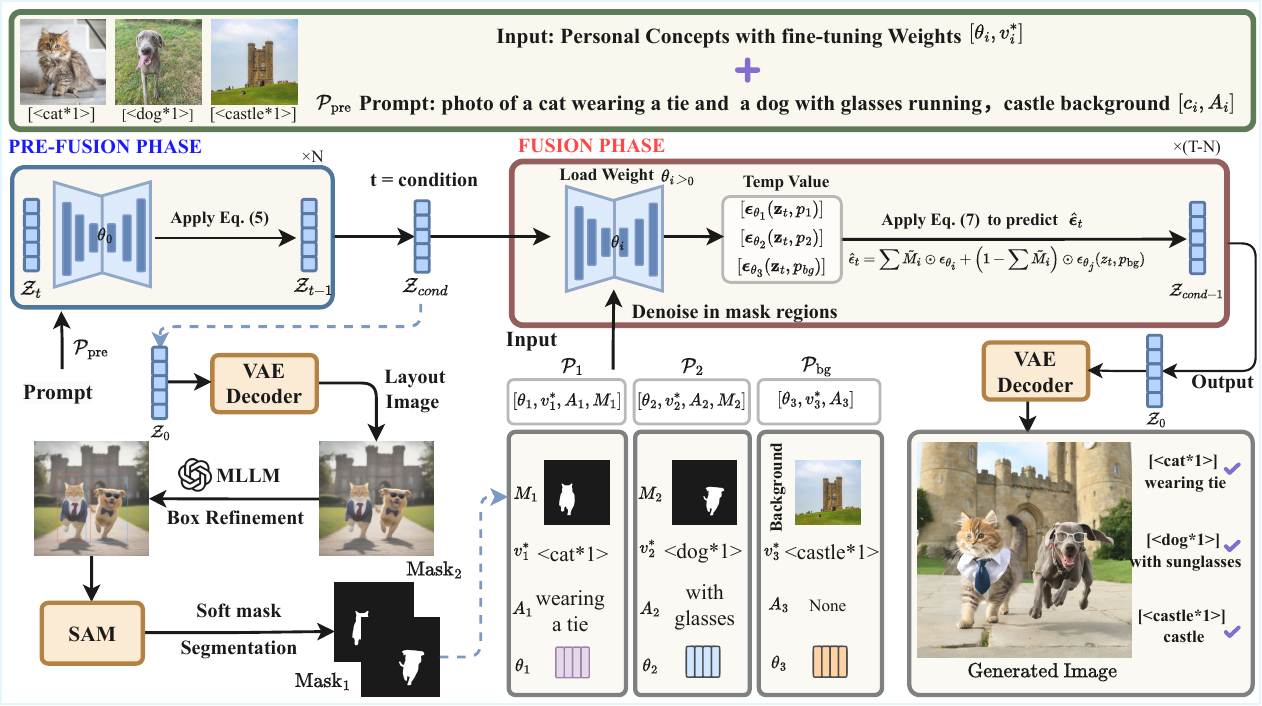}
  \caption{\textbf{Overview of the MultiCompose pipeline.}
  Far left: per-concept reference images and independently fine-tuned weight sets.
  Left: the pre-fusion phase denoises a layout latent under a global
  prompt $\mathcal{P}_{\mathrm{pre}}$ with token-level cross-subject attention isolation
  (Eq.~\ref{eq:prefusion_attn}).
  Center: at the phase boundary $t_{\mathrm{cond}}$, the layout latent
  is decoded, and an MLLM detects per-subject bounding boxes from the
  layout image, which are then processed by a segmentation model to produce
  mutually exclusive soft masks $\{\tilde{M}_i\}$.
  Right: the fusion phase generates per-concept noise predictions under
  single-subject prompts with concept-specific weights, and composes them
  via spatial mask routing (Eq.~\ref{eq:fusion_spatial}) to produce the
  final image. The pre-fusion and fusion phases run for 10 and 40
  denoising steps, respectively.}
  \Description{A pipeline diagram showing global-prompt pre-fusion for
  layout initialization, one-time boundary mask extraction using an MLLM
  and a segmentation model, and per-concept fusion with spatial mask routing.}
  \label{fig:pipeline}
\end{figure*}

The MultiCompose inference pipeline is illustrated in
Fig.~\ref{fig:pipeline}.
Given $N$ independently fine-tuned concept weight sets,
MultiCompose proceeds in two phases:
a pre-fusion phase that establishes the spatial layout of all subjects
under a shared global prompt $\mathcal{P}_{\mathrm{pre}}$ with token-level attention isolation,
followed by a fusion phase that activates per-concept weight sets
under single-subject prompts $\mathcal{P}_{\mathrm{i}}$ and composes the resulting noise predictions
through mutually exclusive soft masks.

\subsubsection{Per-Concept Fine-Tuning with Semantic Preservation}
\label{sec:reg}

Each concept $i$ is independently fine-tuned under the standard denoising objective
(Fig.~\ref{fig:finetune}):
\begin{equation}
  L_{\mathrm{DM}} =
  \mathbb{E}_{\mathbf{z},\boldsymbol{\epsilon},t}
  \bigl\|\boldsymbol{\epsilon}
  - \boldsymbol{\epsilon}_{\theta_i}(\mathbf{z}_t, t,\, \mathcal{P}_i)\bigr\|_2^2,
  \label{eq:ldm}
\end{equation}
where $\mathbf{z}_t$ is the noisy latent at timestep $t$,
$\boldsymbol{\epsilon} \sim \mathcal{N}(0,\mathbf{I})$ is the sampled noise,
and $\mathcal{P}_i$ is the training prompt containing modifier token $\mathbf{v}^*_i$
and its corresponding category word $c_i$
(e.g., ``a photo of a $\mathbf{v}^*_i$ $c_i$'').

As fine-tuning proceeds, $\mathbf{v}^*_i$ drifts away from  the semantic neighborhood
of $c_i$ in the text encoder's representation space~\cite{kumari2023customdiffusion,Tewel2023KeyLockedRO}.
This divergence weakens the cross-attention response of attribute tokens in $A_i$
(the per-subject attribute set defined in Sec.~\ref{sec:formulation})
to the subject's spatial region, thereby degrading attribute binding in downstream composition.
To address this, we introduce a \textbf{semantic preservation regularization} term.
Let $\tau(\cdot)$ denote the EOS-token embedding of a prompt extracted by the frozen CLIP
text encoder~\cite{radford2021learning}.
The regularization term minimizes the cosine distance between the
modifier phrase embedding $\tau(\mathbf{v}^*_i, c_i)$, obtained from the prompt
containing both $\mathbf{v}^*_i$ and $c_i$, and the category phrase embedding
$\tau(c_i)$, obtained from the prompt containing only the category word $c_i$:
\begin{equation}
  L_{\mathrm{sp}} = 1 -
  \frac{\tau(\mathbf{v}^*_i, c_i)\cdot\tau(c_i)}
  {\|\tau(\mathbf{v}^*_i, c_i)\|\,\|\tau(c_i)\|},
  \label{eq:spl}
\end{equation}
and the total fine-tuning objective is:
\begin{equation}
  L = L_{\mathrm{DM}} + \lambda\, L_{\mathrm{sp}},
  \label{eq:total}
\end{equation}
where $\lambda$ is the regularization weight.
The EOS-token embedding aggregates the semantic content of the full modifier
phrase, so this regularization preserves the prior association between subject
identity and attribute composition without requiring additional training data.

\begin{figure}[t]
  \centering
  \includegraphics[width=0.97\linewidth]{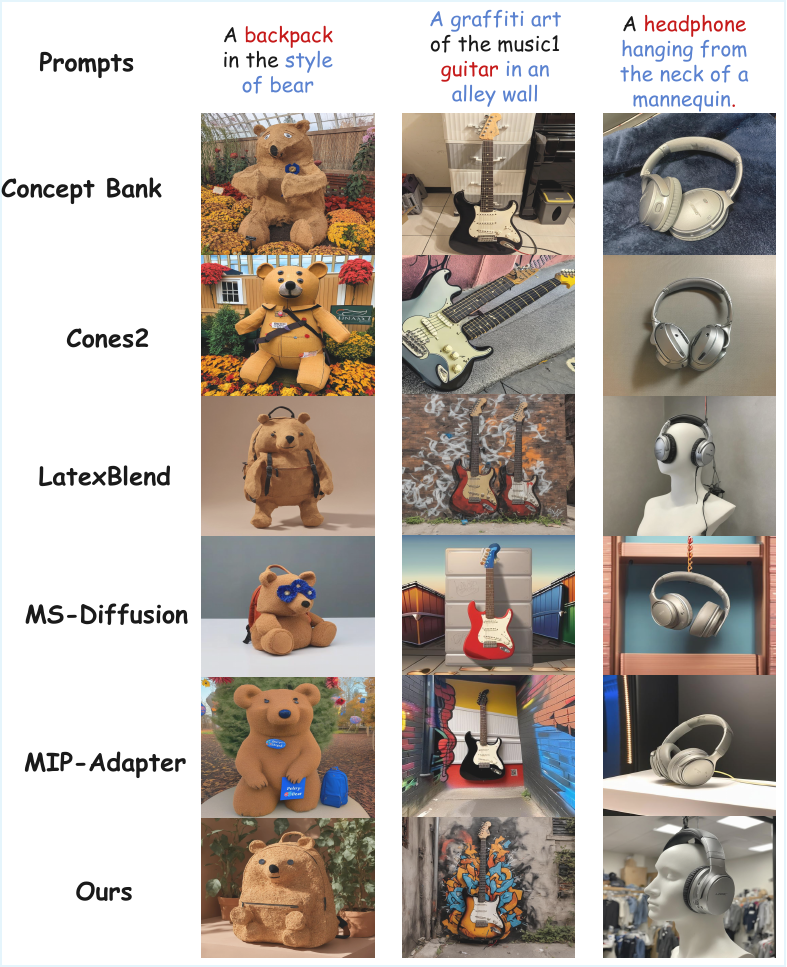}
  \caption{Single-subject personalization with complex attribute binding.
           Each row shows a personalized concept generated under prompts
           involving spatial relationships, style transfer, and
           category-crossing appearance.}
\Description{A grid of single-subject personalized generations illustrating
complex spatial, stylistic, and category-crossing attribute compositions.}
\label{fig:single_attr}
\end{figure}
\subsubsection{Two-Phase Inference Procedure}
\label{sec:inference}
\begin{figure*}[t]
  \centering
  \includegraphics[width=0.94\linewidth]{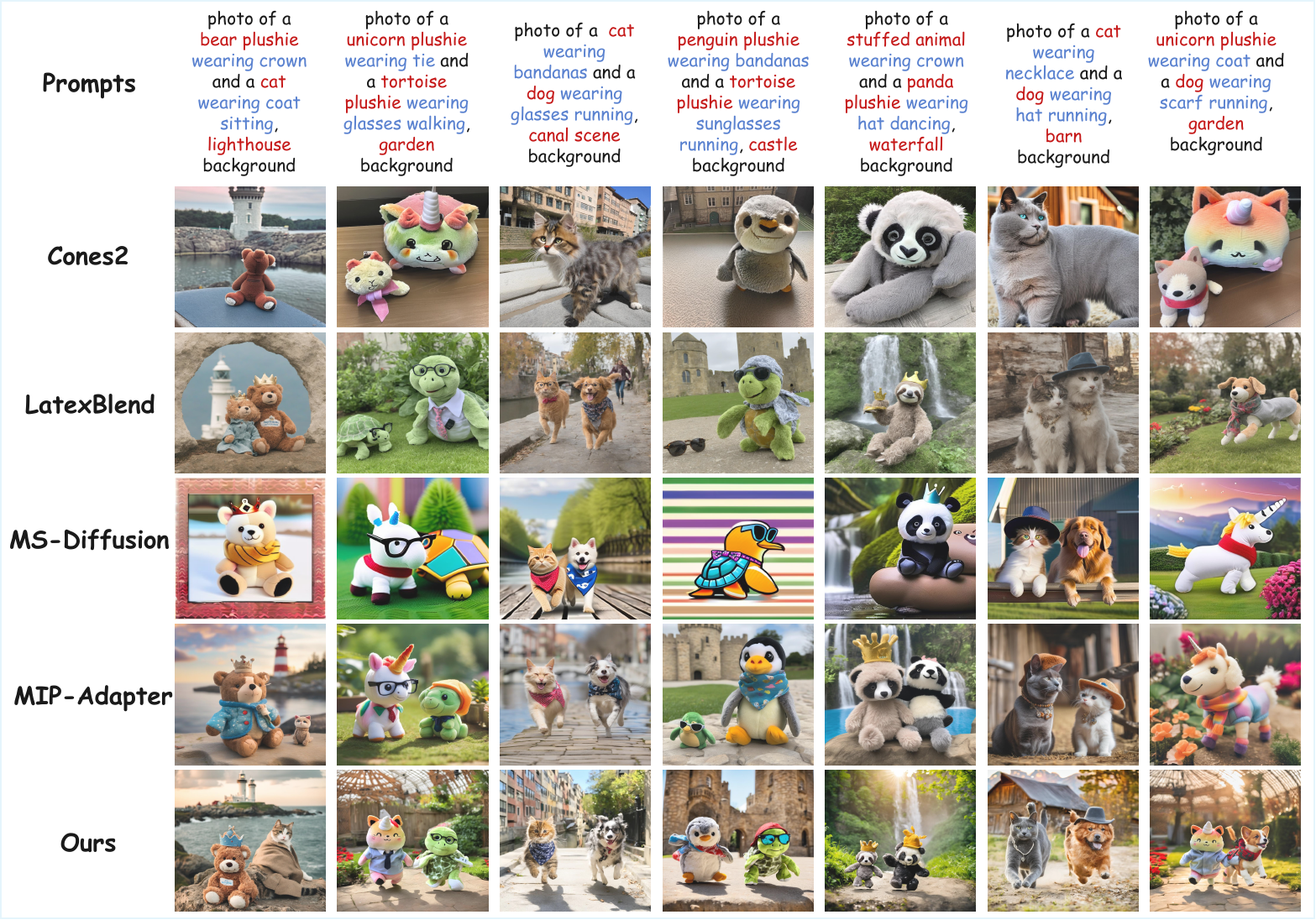}
  \caption{Multi-subject generation with per-subject
           attribute binding.
           Each column shows a different subject combination.
           MultiCompose preserves per-subject identity while correctly
           assigning specified attributes to their designated subjects.}
\Description{A grid of two-subject personalized compositions in which each
subject retains its identity and receives its designated attributes.}
\label{fig:multi_attr}
\end{figure*}

At inference, all $N$ per-concept weight sets are available but activated selectively.
The denoising trajectory is divided into two sequential phases using a
step-count split $t_{\mathrm{cond}}$, defined as a fixed fraction of the
total $T$ denoising steps~\cite{ho2020ddpm,song2020denoising}.
The \emph{pre-fusion} phase comprises the first $t_{\mathrm{cond}}$ steps
and establishes subject layout, while the \emph{fusion} phase comprises
the remaining $T-t_{\mathrm{cond}}$ steps and performs personalized
per-subject generation under spatial constraints.

\medskip
\noindent\textbf{\textit{Pre-Fusion Phase}} (first $t_{\mathrm{cond}}$ steps).\quad
The pre-fusion phase operates on a single global prompt
$\mathcal{P}_{\mathrm{pre}}$ that contains category words $\{c_i\}$ and
attribute descriptions $\{A_i\}$ for all $N$ subjects, with no modifier tokens
$\{\mathbf{v}^*_i\}$.
For example, given two subjects with $c_1 = \texttt{cat}$,
$A_1 = \{\texttt{wearing tie}\}$ and $c_2 = \texttt{dog}$,
$A_2 = \{\texttt{with sunglasses}\}$, the global prompt is
``photo of a cat wearing tie and a dog with sunglasses running, castle background''.
The base model parameters $\theta_0$ are used throughout this phase;
no personalized weight set is activated.
Since $\mathcal{P}_{\mathrm{pre}}$ encodes all subjects in a single
prompt, attribute tokens of one subject may receive high cross-attention
score at spatial positions occupied by another subject, resulting in
attribute leakage.
Following the attribute-object separation principle in~\cite{zhuang2024magnet},
we enforce spatial isolation of attribute attention across subjects.
We first compute a per-subject ownership map from the cross-attention scores
of category tokens:
\begin{equation}
  O_i(\mathbf{s}) =
  \frac{A_{\mathrm{cat}}(\mathbf{s}, c_i)}
       {\sum_{k=1}^{N} A_{\mathrm{cat}}(\mathbf{s}, c_k)},
  \label{eq:ownership}
\end{equation}
where $A_{\mathrm{cat}}(\mathbf{s}, c_i)$ denotes the cross-attention
score between spatial position $\mathbf{s}$ and category token $c_i$.
$O_i(\mathbf{s})$ approaches 1 when position $\mathbf{s}$ is dominated
by subject $i$, and approaches 0 when it belongs to other subjects.
We then suppress the attention logit of each attribute token
$a \in A_i$ at positions where $O_i$ is low:
\begin{equation}
  \tilde{A}(\mathbf{s}, a) =
  \frac{\mathbf{Q}(\mathbf{s})^\top \mathbf{K}(a)}{\sqrt{d}}
  - \alpha\,(1 - O_i(\mathbf{s})),
  \quad a \in A_i,
  \label{eq:prefusion_attn}
\end{equation}
where $\alpha > 0$ controls the suppression strength.
When position $\mathbf{s}$ is owned by subject $i$, the penalty term vanishes
and the original attention logit is preserved;
when $\mathbf{s}$ belongs to other subjects, the negative bias reduces
the resulting attention score, confining attribute $a$ to its designated
subject region.
After the first $t_{\mathrm{cond}}$ steps, the phase yields a layout latent
$\mathbf{z}_{t_{\mathrm{cond}}}$ that encodes the approximate spatial arrangement
of all $N$ subjects with attribute-to-subject associations preserved.

\medskip
\noindent\textbf{\textit{Mask Extraction at the Phase Boundary}.}\quad
The layout latent $\mathbf{z}_{t_{\mathrm{cond}}}$ is decoded to produce a
coarse layout image.
A multimodal large language model (MLLM) is applied to this image to
detect per-subject bounding boxes $\{\hat{b}_i\}$ based on the category words
$\{c_i\}$, without any user-provided spatial prior.
The detected boxes $\{\hat{b}_i\}$ are passed to a segmentation
model~\cite{kirillov2023segment} to generate soft spatial masks $\{M_i\}$.
To enforce spatial mutual exclusion across subjects, masks are normalized at each
spatial position $\mathbf{s}$:
\begin{equation}
  \tilde{M}_i(\mathbf{s}) =
  \frac{M_i(\mathbf{s})}{\sum_{j=1}^{N}M_j(\mathbf{s}) + \varepsilon}.
  \label{eq:mask_norm}
\end{equation}

\medskip
\noindent\textbf{\textit{Fusion Phase}} (remaining $T-t_{\mathrm{cond}}$ steps).\quad
In the fusion phase (right portion of Fig.~\ref{fig:pipeline}),
each subject $i$ is denoised under its own single-subject prompt $\mathcal{P}_i$,
which contains the modifier token $\mathbf{v}^*_i$ and the attribute
descriptions in $A_i$, using the concept-specific weight set $\theta_i$.
Each prompt $\mathcal{P}_i$ describes only subject $i$ and its
attributes; no information about other subjects is present in the text conditioning~\cite{ruiz2023dreambooth,kumari2023customdiffusion}.
Each forward pass through the U-Net is therefore executed with a single concept's
weight set and a single-subject prompt, so cross-attention operates exclusively
over the tokens of subject $i$.
The per-concept noise predictions are then composed within their respective mask
regions, with a background-specific personalized concept weight $\theta_j$ ($j \notin \{1,\dots,N\}$)
and prompt $\mathcal{P}_{\mathrm{bg}}$ handling regions outside all subject masks:
\begin{equation}
  \hat{\boldsymbol{\epsilon}}_t =
  \sum_{i=1}^{N}\tilde{M}_i \odot
  \boldsymbol{\epsilon}_{\theta_i}(\mathbf{z}_t, \mathcal{P}_i)
  + \Bigl(1-\textstyle\sum_{i=1}^N\tilde{M}_i\Bigr)\odot
  \boldsymbol{\epsilon}_{\theta_j}(\mathbf{z}_t, \mathcal{P}_{\mathrm{bg}}).
  \label{eq:fusion_spatial}
\end{equation}
The mutually exclusive masks $\tilde{M}_i$ ensure that each concept's
noise prediction contributes only to its designated spatial region,
eliminating cross-subject attribute leakage during the fusion phase.

\subsection{MSP-Bench: Multi-Subject Personalization Benchmark}
\label{sec:bench}

No existing benchmark jointly evaluates identity fidelity, attribute
binding, and attribute misalignment in the personalized multi-subject setting~\cite{peng2024dreambench,ku2024viescore}.
We introduce \textbf{MSP-Bench}, which scores each of these three dimensions
through a dual-pathway protocol: a \emph{traditional pathway}~($h$) based on
visual feature similarity and an \emph{MLLM pathway}~($m$) based on
multimodal language model judgments~\cite{hurst2024gpt,zhang2023gpt}, combined as
$d = w_h\,d_h + w_m\,d_m$ ($w_h{+}w_m{=}1$).

\smallskip
\noindent\textbf{\textit{Identity Fidelity (ID).}}\quad
ID measures how faithfully each personalized subject preserves its reference
identity.
In the traditional pathway, each subject region is cropped via
mask~$\tilde{M}_i$ and compared against reference set~$R_i$ using
DINO~\cite{caron2021dino} cosine similarity.
In the MLLM pathway, the cropped subject region and each reference image
are provided to a multimodal language model, which outputs a similarity
judgment on a continuous scale.
When a personalized background is present, its identity score is incorporated
with weight~$w_{\mathrm{bg}}{=}0.5$, reflecting its lower perceptual salience
relative to foreground subjects:
\begin{equation}
  \mathrm{ID} =
  \frac{\displaystyle\sum_{i=1}^{N}\mathrm{ID}_i
        + w_{\mathrm{bg}}\,\mathrm{ID}_{\mathrm{bg}}}
       {N + w_{\mathrm{bg}}},
  \label{eq:id}
\end{equation}
where $\mathrm{ID}_i{=}\frac{1}{|R_i|}\sum_{r\in R_i}
  \cos(f(\mathrm{crop}(\hat{x},\tilde{M}_i)),f(r))$
for the traditional pathway, $f(\cdot)$ denotes the DINO feature extractor,
and $N$ is the number of foreground subjects.

\smallskip
\noindent\textbf{\textit{Attribute Binding Accuracy (BIND).}}\quad
BIND measures whether each specified attribute is correctly presented on its
designated subject.
In the traditional pathway, CLIP~\cite{radford2021learning,hessel2021clipscore}-based feature matching is used to assess
whether the attribute description aligns with the visual content of the
subject crop.
In the MLLM pathway, the cropped region and the attribute description are
provided to a multimodal language model, which returns a binary correctness
judgment:
\begin{equation}
  \mathrm{BIND}_i =
  \frac{1}{|A_i|}\sum_{a\in A_i}
  \mathbf{1}\!\bigl[\,\phi\!\bigl(\mathrm{crop}(\hat{x},\tilde{M}_i),\, a\bigr)=1\,\bigr],
  \label{eq:bind}
\end{equation}
where $\phi(\cdot,\cdot)$ is a binary classifier instantiated by either pathway.
BIND rewards correct binding only; error penalization is handled by MIS.

\smallskip
\noindent\textbf{\textit{Attribute Misalignment (MIS).}}\quad
MIS captures three distinct failure modes that BIND alone cannot distinguish.
For each subject~$i$, the evaluation procedure examines every attribute
$a \in A_i$ to detect the following errors:
\begin{itemize}
  \item \emph{Leakage} ($\mathrm{Leak}_i$): For each pair $(i,j)$ with
        $i{\neq}j$, the MLLM is queried whether attribute~$a$ of subject~$i$
        appears in the cropped region of subject~$j$.
        The traditional pathway computes CLIP similarity between the attribute
        description and the crop of subject~$j$.
        $\mathrm{Leak}_i$ is the fraction of attributes that are detected
        in at least one other subject's region.
  \item \emph{Confusion} ($\mathrm{Conf}_i$): The MLLM is queried whether
        the attribute present on subject~$i$ matches the specified description
        or has been replaced by a semantically similar but incorrect one
        (e.g., \emph{coat}~$\to$~\emph{jacket}).
        This error type relies primarily on the MLLM pathway, as
        distinguishing fine-grained semantic substitutions is beyond the
        capability of feature-level matching.
        $\mathrm{Conf}_i$ is the fraction of attributes judged as confused.
  \item \emph{Missing} ($\mathrm{Miss}_i$): An attribute is scored as missing
        if neither the traditional nor the MLLM pathway detects its presence
        in subject~$i$'s region.
        $\mathrm{Miss}_i$ is the fraction of entirely absent attributes.
\end{itemize}
These sub-scores are aggregated with severity-based weights:
\begin{equation}
  \mathrm{MIS}_i = 0.5\,\mathrm{Leak}_i
                  + 0.35\,\mathrm{Conf}_i
                  + 0.15\,\mathrm{Miss}_i.
  \label{eq:mis}
\end{equation}
Leakage receives the highest weight because it corrupts multiple
subject regions simultaneously and typically co-occurs with binding failures.
Lower MIS indicates fewer attribute errors.

\smallskip
\noindent\textbf{\textit{MSP Score.}}\quad
The three dimensions are aggregated into a single score.
Since MIS is a penalty (lower is better), it enters as $1{-}\mathrm{MIS}$:
\begin{equation}
  \mathrm{MSP} = 0.50\,\mathrm{ID}
                  + 0.35\,\mathrm{BIND}
                  + 0.15\,(1-\mathrm{MIS}).
  \label{eq:final}
\end{equation}

\smallskip
\noindent\textbf{\textit{Benchmark Construction.}}\quad
MSP-Bench covers attribute categories including color, material, worn items,
and held objects, with two to three subjects per prompt.
Per-concept reference images are drawn from standard personalization
evaluation sets.
All metrics are computed automatically without human annotation,
enabling reproducible large-scale evaluation.

\section{Experiments}

\subsection{Implementation Details}

\begin{table}[t]
  \centering
  \caption{Quantitative comparison on personalized single-subject
           generation on the Concept101 dataset~\cite{kumari2023customdiffusion}.
           $\uparrow$: higher is better.
           Best results in \textbf{bold}, second best \underline{underlined}.}
  \label{tab:quantitative}
  \begin{tabular}{lccc}
    \hline
    Method & CLIP-T$\uparrow$ & CLIP-I$\uparrow$ & DINO$\uparrow$ \\
    \hline
    Textual Inversion~\cite{gal2022textual}    & 0.6117 & 0.7530 & 0.5128 \\
    DreamBooth~\cite{ruiz2023dreambooth}        & 0.7514 & 0.7521 & 0.5541 \\
    Custom Diffusion~\cite{kumari2023customdiffusion} & 0.7583 & 0.7456 & 0.5335 \\
    Custom Diffusion (SDXL)                     & 0.7537 & \underline{0.7769} & 0.5382   \\
    MIP-Adapter~\cite{huang2025resolving}                     &\underline{0.7586} &\textbf{0.7877} &\textbf{0.5771} \\
    \hline
    MultiCompose (Ours)  & \textbf{0.8164} & 0.7671 & \underline{0.5614} \\
    \hline
  \end{tabular}
\end{table}

All experiments use Stable Diffusion XL~\cite{podell2023sdxl} as the base model
unless otherwise noted.
Per-concept fine-tuning follows the cross-attention key--value update strategy
described in Section~\ref{sec:reg}, with learning rate $10^{-5}$,
batch size 2, and 1000 optimization steps per concept.
Each concept uses 3--15 reference images for training.
The semantic preservation weight is set to $\lambda=1.0$ throughout.
The denoising trajectory uses 50 DDIM steps with guidance scale 6.0.
We set the step-count split to $t_{\mathrm{cond}}=0.2T=10$:
the first 10 denoising steps form the pre-fusion phase, and the remaining
$T-t_{\mathrm{cond}}=0.8T=40$ steps form the fusion phase.
Soft masks are generated by prompting SAM~\cite{kirillov2023segment} with
MLLM-detected bounding boxes obtained from the layout image
decoded once at the phase boundary.
The cross-subject attention suppression weight is set to $\alpha = 0.1$ in all experiments.
All experiments are conducted on 2$\times$ RTX 4090 GPUs.

\subsection{Baselines}

We compare MultiCompose with \textbf{Textual Inversion}~\cite{gal2022textual},
\textbf{DreamBooth}~\cite{ruiz2023dreambooth},
\textbf{Custom Diffusion}~\cite{kumari2023customdiffusion}, and
\textbf{MIP-Adapter}~\cite{huang2025resolving}.
Additional Concept101 baselines are
\textbf{Mix-of-Show}~\cite{gu2024mixofshow} and
\textbf{MC$^2$}~\cite{jiang2024mc2}.
All applicable methods use 3--15 training images per concept.

\subsection{Quantitative Evaluation}

MultiCompose achieves the highest CLIP-T score among all methods,
while ranking second on DINO behind MIP-Adapter.
Compared to Custom Diffusion (SDXL) on the same backbone,
MultiCompose improves CLIP-T by 0.063,
indicating that the two-phase inference with spatial mask routing
improves text-image alignment beyond per-concept fine-tuning alone.

On Concept101 (Table~\ref{tab:concept101}), MultiCompose achieves the
best CLIP-T (0.7982; $+$0.023 over MIP-Adapter) and ranks second on
CLIP-I and DINO, while adding no trainable parameters beyond
per-concept fine-tuning.

\begin{table}[t]
  \centering
  \caption{Quantitative comparison on the Concept101
           multi-concept benchmark~\cite{kumari2023customdiffusion}.
           Each concept uses 3--15 reference images.
           $\uparrow$: higher is better.
           Best results in \textbf{bold}, second best \underline{underlined}.
           $\S$: results reported in the cited papers.}
  \label{tab:concept101}
  \begin{tabular}{lccc}
    \hline
    Method & CLIP-T$\uparrow$ & CLIP-I$\uparrow$ & DINO$\uparrow$ \\
    \hline
    DreamBooth~\cite{ruiz2023dreambooth}                        & 0.7383 & 0.6636 & 0.3849 \\
    Custom Diffusion (Opt)~\cite{kumari2023customdiffusion}     & 0.7599 & 0.6595 & 0.3684 \\
    Custom Diffusion (Joint)~\cite{kumari2023customdiffusion}   & 0.7534 & 0.6704 & 0.3799 \\
    Mix-of-Show\textsuperscript{\S}~\cite{gu2024mixofshow}                  & 0.7280 & 0.6700 & 0.3940 \\
    MC$^{2}$\textsuperscript{\S}~\cite{jiang2024mc2}                          & 0.7670 & 0.6860 & 0.4060 \\
    MIP-Adapter (SDXL)~\cite{huang2025resolving}                 & \underline{0.7750} & \textbf{0.6950} & \textbf{0.4397} \\
    \hline
    MultiCompose (Ours, SDXL)  & \textbf{0.7982} & \underline{0.6887} & \underline{0.4282} \\
    \hline
  \end{tabular}
\end{table}

On MSP-Bench (Table~\ref{tab:mspbench}), MultiCompose achieves the
best MSP (0.6932; $+$0.105 over MIP-Adapter) and BIND ($+$0.130),
and the lowest MIS (0.3922), demonstrating improved binding with
fewer leakage, confusion, and missing errors.

\begin{figure*}[t]
  \centering
  \includegraphics[width=0.87\linewidth]{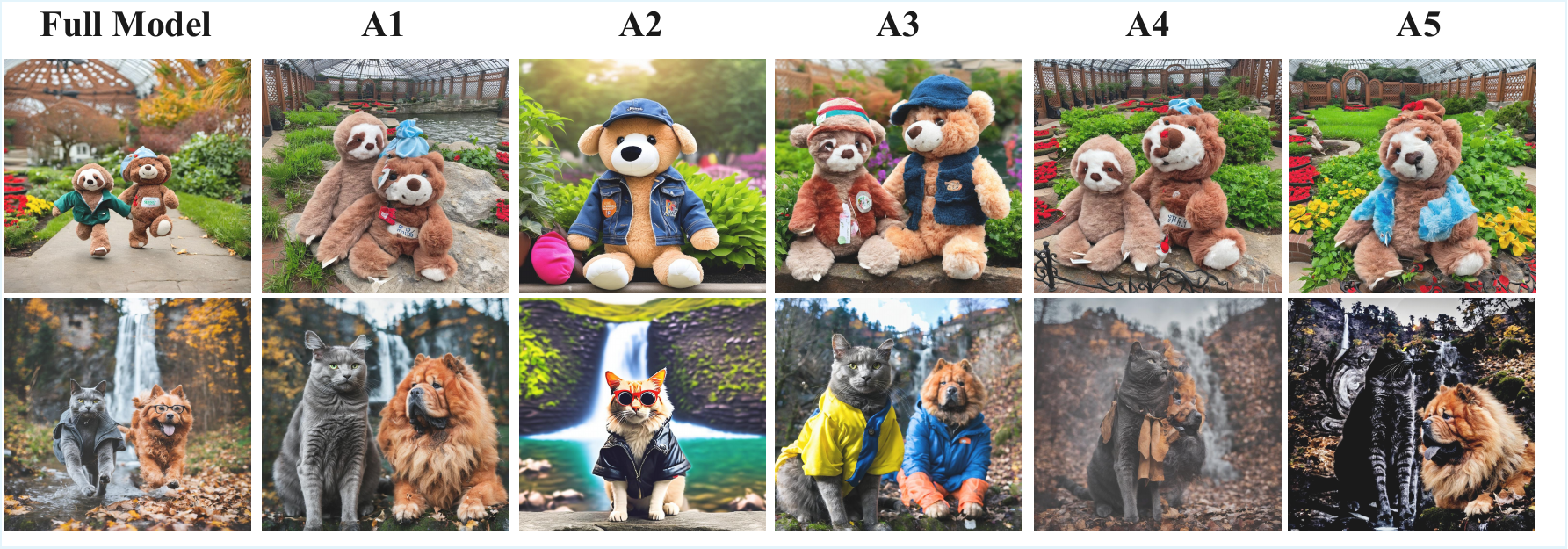}
  \caption{Visual comparison of ablation variants (A1)--(A5).
           Each column removes one component from the full pipeline.
           Failure modes include attribute loss, layout degradation,
           attribute leakage, boundary mixing, and attribute entanglement.}
  \Description{A grid comparing the full MultiCompose model with five
  ablations. The ablations illustrate attribute loss, layout degradation,
  attribute leakage, boundary mixing, and attribute entanglement.}
  \label{fig:ablation}
\end{figure*}

\subsection{Qualitative Results}

Fig.~\ref{fig:single_attr} shows that MultiCompose preserves personalized
identity and binds complex spatial, stylistic, and category-crossing
attributes to a single subject.
For multi-subject composition (Fig.~\ref{fig:multi_attr}), baselines exhibit
attribute leakage or identity degradation, whereas MultiCompose preserves
distinct identities and assigns each attribute to its designated subject
through spatial mask routing.

\begin{table}[t]
  \centering
  \caption{Comparison on MSP-Bench.
           ID and BIND: higher is better ($\uparrow$).
           MIS: lower is better ($\downarrow$).
           MSP is the weighted aggregate (Eq.~\ref{eq:final}).
           Best results in \textbf{bold}, second best \underline{underlined}.}
  \label{tab:mspbench}
  \begin{tabular}{lcccc}
    \hline
    Method & ID$\uparrow$ & BIND$\uparrow$ & MIS$\downarrow$ & MSP$\uparrow$ \\
    \hline
    Cones 2~\cite{liu2023cones2}       & 0.6913 & 0.2321 & 0.5963 & 0.4352 \\
    MS-Diffusion~\cite{wang2025msdiffusion}  & 0.5755 & 0.4586 & 0.5809 & 0.5069 \\
    LatentBlend~\cite{jin2025latexblend}   & 0.7182 & 0.4571 & 0.5581 & 0.5506 \\
    MIP-Adapter   & \underline{0.7184} & \underline{0.4969} & \underline{0.5500} & \underline{0.5886} \\
    \hline
    MultiCompose (Ours) & \textbf{0.7735} & \textbf{0.6271} & \textbf{0.3922} & \textbf{0.6932} \\
    \hline
  \end{tabular}
\end{table}

\begin{table*}[ht]
  \centering
  \caption{Ablation results.
           Left: MSP-Bench metrics. Right: conventional metrics.
           Best per column in \textbf{bold}.}
  \label{tab:ablation}
  \begin{tabular}{clcccc|ccc}
    \hline
    {} & Variant
      & ID$\uparrow$ & BIND$\uparrow$
      & MIS$\downarrow$ & MSP$\uparrow$
      & CLIP-I$\uparrow$ & CLIP-T$\uparrow$
      & DINO$\uparrow$ \\
    \hline
    (A1) & w/o $L_{\mathrm{sp}}$
      & \textbf{0.7893} & 0.4442
      & 0.4869 & 0.6050
      & \textbf{0.7113} & 0.8000
      & \textbf{0.4671} \\
    (A2) & w/o pre-fusion
      & 0.6179 & 0.5842
      & 0.5856 & 0.5544
      & 0.6719 & \textbf{0.9116}
      & 0.3399 \\
    (A3) & w/o attn.\ suppression
      & 0.7728 & 0.5960
      & 0.6482 & 0.6271
      & 0.6665 & 0.9082
      & 0.4257 \\
    (A4) & w/o mask exclusivity
      & 0.7308 & 0.4563
      & 0.5050 & 0.5846
      & 0.6839 & 0.7988
      & 0.4032 \\
   (A5) & Shared-prompt fusion
      & 0.6460 & 0.4972
      & 0.6050 & 0.5475
      & 0.6864 & 0.8069
      & 0.4208 \\
    \hline
    & Full MultiCompose
      & 0.7735 & \textbf{0.6271}
      & \textbf{0.3922} & \textbf{0.6932}
      & 0.6665 & 0.8997
      & 0.3798 \\
    \hline
  \end{tabular}
\end{table*}
\subsection{Ablation Study}

We evaluate five component-wise ablations using MSP-Bench and conventional
metrics (Table~\ref{tab:ablation}); Fig.~\ref{fig:ablation} shows their
corresponding failure modes.

\noindent\textbf{(A1) w/o $L_{\mathrm{sp}}$.}\quad
Without semantic preservation, the modifier token diverges from its
category word.
Although ID remains high (0.7893), BIND drops by 0.183 and MIS rises by 0.095.
Moreover, (A1) obtains the highest CLIP-I (0.711) and DINO (0.467), showing
that conventional metrics can overlook missing subject attributes.

\noindent\textbf{(A2) w/o pre-fusion.}\quad
Without the initial $t_{\mathrm{cond}}$-step pre-fusion, masks are derived
directly from text and the output collapses to a single subject, causing
an MSP drop of 0.139 and the largest ID drop ($-$0.156); MIS rises to
0.586 despite the highest CLIP-T (0.912).

\noindent\textbf{(A3) w/o attn.\ suppression.}\quad
Without cross-subject attention suppression, attributes leak across
subject boundaries; MIS rises to 0.648 although ID (0.773) and CLIP-T
(0.908) remain high.

\noindent\textbf{(A4) w/o mask exclusivity.}\quad
Without mask exclusivity, overlapping concept predictions mix identities
at subject boundaries, reducing ID to 0.731 and increasing MIS to 0.505.

\noindent\textbf{(A5) Shared-prompt fusion.}\quad
Using the shared global prompt during fusion causes identity degradation
and attribute entanglement because all subject tokens compete in
cross-attention despite mask routing, yielding the lowest MSP (0.548).

Together, these ablations show that MSP-Bench captures failures overlooked
by conventional metrics.

\section{Conclusion}

We presented MultiCompose, a framework that composes multiple
independently personalized subjects with per-subject attribute binding
through semantic preservation regularization and two-phase spatial
inference.
Experimental results demonstrate that MultiCompose achieves superior
performance in multi-subject personalized generation, maintaining
high identity fidelity while correctly binding attributes to their
designated subjects.
We further introduced MSP-Bench, an effective benchmark for joint
evaluation of identity fidelity, attribute binding, and attribute
misalignment in the personalized multi-subject setting.
The benchmark provides assessment of failure modes that
conventional metrics overlook, enabling systematic evaluation of
future methods in this domain.

\begin{acks}
This work was partially supported by the National Natural Science Foundation of China (No.~62272227).
\end{acks}

\bibliographystyle{ACM-Reference-Format}
\bibliography{references}

@inproceedings{rombach2022ldm,
  title     = {High-Resolution Image Synthesis with Latent Diffusion Models},
  author    = {Rombach, Robin and Blattmann, Andreas and Lorenz, Dominik and Esser, Patrick and Ommer, Bj{\"o}rn},
  booktitle = {Proceedings of the IEEE/CVF Conference on Computer Vision and Pattern Recognition (CVPR)},
  pages     = {10684--10695},
  year      = {2022}
}

@article{podell2023sdxl,
  title   = {{SDXL}: Improving Latent Diffusion Models for High-Resolution Image Synthesis},
  author  = {Podell, Dustin and English, Zion and Lacey, Kyle and Blattmann, Andreas and Dockhorn, Tim and M{\"u}ller, Jonas and Penna, Joe and Rombach, Robin},
  journal = {arXiv preprint arXiv:2307.01952},
  year    = {2023}
}

@inproceedings{ruiz2023dreambooth,
  title     = {{DreamBooth}: Fine Tuning Text-to-Image Diffusion Models for Subject-Driven Generation},
  author    = {Ruiz, Nataniel and Li, Yuanzhen and Jampani, Varun and Pritch, Yael and Rubinstein, Michael and Aberman, Kfir},
  booktitle = {Proceedings of the IEEE/CVF Conference on Computer Vision and Pattern Recognition (CVPR)},
  pages     = {22500--22510},
  year      = {2023}
}

@inproceedings{gal2022textual,
  title     = {An Image is Worth One Word: Personalizing Text-to-Image Generation using Textual Inversion},
  author    = {Gal, Rinon and Alaluf, Yuval and Atzmon, Yuval and Patashnik, Or and Bermano, Amit Haim and Chechik, Gal and Cohen-Or, Daniel},
  booktitle = {International Conference on Learning Representations (ICLR)},
  year      = {2023}
}

@inproceedings{hu2022lora,
  title     = {{LoRA}: Low-Rank Adaptation of Large Language Models},
  author    = {Hu, Edward J. and Shen, Yelong and Wallis, Phillip and Allen-Zhu, Zeyuan and Li, Yuanzhi and Wang, Shean and Wang, Lu and Chen, Weizhu},
  booktitle = {International Conference on Learning Representations (ICLR)},
  year      = {2022}
}

@inproceedings{gu2024mixofshow,
  title     = {Mix-of-Show: Decentralized Low-Rank Adaptation for Multi-Concept Customization of Diffusion Models},
  author    = {Gu, Yuchao and Wang, Xintao and Wu, Jay Zhangjie and Shi, Yujun and Chen, Yunpeng and Fan, Zihan and Xiao, Wuyou and Zhao, Rui and Chang, Shuning and Wu, Weijia and Ge, Yixiao and Shan, Ying and Shou, Mike Zheng},
  booktitle = {Advances in Neural Information Processing Systems (NeurIPS)},
  year      = {2024}
}

@inproceedings{bar2023multidiffusion,
  title     = {{MultiDiffusion}: Fusing Diffusion Paths for Controlled Image Generation},
  author    = {Bar-Tal, Omer and Yariv, Lior and Lipman, Yaron and Dekel, Tali},
  booktitle = {Proceedings of the 40th International Conference on Machine Learning (ICML)},
  pages     = {1737--1752},
  year      = {2023}
}

@inproceedings{avrahami2023spatext,
  title     = {{SpaText}: Spatio-Textual Representation for Controllable Image Generation},
  author    = {Avrahami, Omri and Hayes, Thomas and Gafni, Oran and Gupta, Sonal and Taigman, Yaniv and Parikh, Devi and Lischinski, Dani and Fried, Ohad and Yin, Xi},
  booktitle = {Proceedings of the IEEE/CVF Conference on Computer Vision and Pattern Recognition (CVPR)},
  pages     = {18370--18380},
  year      = {2023}
}

@inproceedings{kirillov2023segment,
  title     = {Segment Anything},
  author    = {Kirillov, Alexander and Mintun, Eric and Ravi, Nikhila and Mao, Hanzi and
               Rolland, Chloe and Gustafson, Laura and Xiao, Tete and Whitehead, Spencer
               and Berg, Alexander C. and Lo, Wan-Yen and Doll{\'a}r, Piotr and
               Girshick, Ross},
  booktitle = {Proceedings of the IEEE/CVF International Conference on Computer Vision (ICCV)},
  pages     = {4015--4026},
  year      = {2023}
}

@inproceedings{caron2021dino,
  title     = {Emerging Properties in Self-Supervised Vision Transformers},
  author    = {Caron, Mathilde and Touvron, Hugo and Misra, Ishan and J{\'e}gou, Herv{\'e}
               and Mairal, Julien and Bojanowski, Piotr and Joulin, Armand},
  booktitle = {Proceedings of the IEEE/CVF International Conference on Computer Vision (ICCV)},
  pages     = {9650--9660},
  year      = {2021}
}

@inproceedings{kumari2023customdiffusion,
  title     = {Multi-Concept Customization of Text-to-Image Diffusion},
  author    = {Kumari, Nupur and Zhang, Bingliang and Zhang, Richard and Shechtman, Eli and Zhu, Jun-Yan},
  booktitle = {Proceedings of the IEEE/CVF Conference on Computer Vision and Pattern Recognition (CVPR)},
  pages     = {1931--1941},
  year      = {2023}
}

@inproceedings{liu2023cones2,
  title =        {Cones 2: Customizable Image Synthesis with Multiple Subjects},
  author =       {Zhiheng Liu and Yifei Zhang and Yujun Shen and Kecheng Zheng and Kai Zhu and Ruili Feng and Yu Liu and Deli Zhao and Jingren Zhou and Yang Cao},
  booktitle =    {Advances in Neural Information Processing Systems (NeurIPS)},
  pages =        {57500--57519},
  year =         {2023}
}

@article{peng2024dreambench,
  title   = {{DreamBench++}: A Human-Aligned Benchmark for Personalized Image Generation},
  author  = {Peng, Yuang and Zhao, Yuxin and Wei, Haoxin and Chen, Zhangye and Xiao, Wenzhao and Yao, Jiwen and Li, Zhiyuan and Liu, Yichen and Chen, Weiming and Zhao, Quanlong and Zhang, Jing and Han, Hu and Dong, Hao},
  journal = {arXiv preprint arXiv:2406.16855},
  year    = {2024}
}

@article{huang2023t2icompbench,
  title   = {{T2I-CompBench++}: An Enhanced and Comprehensive Benchmark for Compositional Text-to-Image Generation},
  author  = {Huang, Kaiyi and Sun, Kaiyue and Xie, Enze and Li, Zhenguo and Liu, Xihui},
  journal = {arXiv preprint arXiv:2307.06350},
  year    = {2023}
}

@inproceedings{ho2020ddpm,
  author    = {Jonathan Ho and Ajay Jain and Pieter Abbeel},
  title     = {Denoising Diffusion Probabilistic Models},
  booktitle = {Advances in Neural Information Processing Systems (NeurIPS)},
  volume    = {33},
  pages     = {6840--6851},
  year      = {2020}
}

@article{song2020denoising,
  title={Denoising Diffusion Implicit Models},
  author={Song, Jiaming and Meng, Chenlin and Ermon, Stefano},
  journal={arXiv:2010.02502},
  year={2020},
  month={October},
  abbr={Preprint},
  url={https://arxiv.org/abs/2010.02502}
}

@inproceedings{hertz2022prompt,
  author    = {Amir Hertz and Ron Mokady and Jay Tenenbaum and
               Kfir Aberman and Yael Pritch and Daniel Cohen-Or},
  title     = {Prompt-to-Prompt Image Editing with Cross Attention Control},
  booktitle = {International Conference on Learning Representations (ICLR)},
  year      = {2023}
}

@article{chefer2023attend,
  author    = {Hila Chefer and Yuval Alaluf and Yael Vinker and
               Lior Wolf and Daniel Cohen-Or},
  title     = {Attend-and-Excite: Attention-Based Semantic Guidance
               for Text-to-Image Diffusion Models},
  journal   = {ACM Transactions on Graphics},
  volume    = {42},
  number    = {4},
  pages     = {148},
  year      = {2023}
}

@inproceedings{kwon2024tweediemix,
  author    = {Gihyun Kwon and Jong Chul Ye},
  title     = {{TweedieMix}: Improving Multi-Concept Fusion for
               Diffusion-based Image/Video Generation},
  booktitle = {International Conference on Learning Representations (ICLR)},
  year      = {2025}
}

@article{jiang2024mc2,
  title   = {{MC$^2$}: Multi-Concept Guidance for Customized Multi-Concept Generation},
  author  = {Jiang, Jiaxiu and Gao, Yabo and Ye, Qingxu and Liu, Xiao and Wang, Jincheng and Zheng, Xian and Tai, Ying},
  journal = {arXiv preprint arXiv:2404.05268},
  year    = {2024}
}

@inproceedings{zhuang2024magnet,
  title     = {Magnet: We Never Know How Text-to-Image Diffusion Models Work,
               Until We Learn How Vision-Language Models Function},
  author    = {Zhuang, Chenyi and Hu, Ying and Gao, Pan},
  booktitle = {Advances in Neural Information Processing Systems (NeurIPS)},
  year      = {2024}
}

@inproceedings{huang2025resolving,
  title={Resolving multi-condition confusion for finetuning-free personalized image generation},
  author={Huang, Qihan and Fu, Siming and Liu, Jinlong and Jiang, Hao and Yu, Yipeng and Song, Jie},
  booktitle={Proceedings of the AAAI conference on Artificial Intelligence},
  volume={39},
  number={4},
  pages={3707--3714},
  year={2025}
}

@inproceedings{ku2024viescore,
  title={Viescore: Towards explainable metrics for conditional image synthesis evaluation},
  author={Ku, Max and Jiang, Dongfu and Wei, Cong and Yue, Xiang and Chen, Wenhu},
  booktitle={Proceedings of the 62nd Annual Meeting of the Association for Computational Linguistics (Volume 1: Long Papers)},
  pages={12268--12290},
  year={2024}
}

@inproceedings{hessel2021clipscore,
  title={Clipscore: A reference-free evaluation metric for image captioning},
  author={Hessel, Jack and Holtzman, Ari and Forbes, Maxwell and Le Bras, Ronan and Choi, Yejin},
  booktitle={Proceedings of the 2021 conference on empirical methods in natural language processing},
  pages={7514--7528},
  year={2021}
}

@article{zhang2023gpt,
  title={Gpt-4v (ision) as a generalist evaluator for vision-language tasks},
  author={Zhang, Xinlu and Lu, Yujie and Wang, Weizhi and Yan, An and Yan, Jun and Qin, Lianke and Wang, Heng and Yan, Xifeng and Wang, William Yang and Petzold, Linda Ruth},
  journal={arXiv preprint arXiv:2311.01361},
  year={2023}
}

@article{hurst2024gpt,
  title={Gpt-4o system card},
  author={Hurst, Aaron and Lerer, Adam and Goucher, Adam P and Perelman, Adam and Ramesh, Aditya and Clark, Aidan and Ostrow, AJ and Welihinda, Akila and Hayes, Alan and Radford, Alec and others},
  journal={arXiv preprint arXiv:2410.21276},
  year={2024}
}

@article{saharia2022photorealistic,
  title={Photorealistic text-to-image diffusion models with deep language understanding},
  author={Saharia, Chitwan and Chan, William and Saxena, Saurabh and Li, Lala and Whang, Jay and Denton, Emily L and Ghasemipour, Kamyar and Gontijo Lopes, Raphael and Karagol Ayan, Burcu and Salimans, Tim and others},
  journal={Advances in neural information processing systems},
  volume={35},
  pages={36479--36494},
  year={2022}
}

@inproceedings{feng2023trainingfree,
    title={Training-Free Structured Diffusion Guidance for Compositional Text-to-Image Synthesis},
    author={Weixi Feng and Xuehai He and Tsu-Jui Fu and Varun Jampani and Arjun Reddy Akula and Pradyumna Narayana and Sugato Basu and Xin Eric Wang and William Yang Wang},
    booktitle={The Eleventh International Conference on Learning Representations },
    year={2023},
    url={https://openreview.net/forum?id=PUIqjT4rzq7}
}

@misc{zhang2023adding,
  title={Adding Conditional Control to Text-to-Image Diffusion Models}, 
  author={Lvmin Zhang and Anyi Rao and Maneesh Agrawala},
  booktitle={IEEE International Conference on Computer Vision (ICCV)},
  year={2023},
}

@article{li2023gligen,
  title={GLIGEN: Open-Set Grounded Text-to-Image Generation},
  author={Li, Yuheng and Liu, Haotian and Wu, Qingyang and Mu, Fangzhou and Yang, Jianwei and Gao, Jianfeng and Li, Chunyuan and Lee, Yong Jae},
  journal={CVPR},
  year={2023}
}

@inproceedings{han2023svdiff,
  title={Svdiff: Compact parameter space for diffusion fine-tuning},
  author={Han, Ligong and Li, Yinxiao and Zhang, Han and Milanfar, Peyman and Metaxas, Dimitris and Yang, Feng},
  booktitle={Proceedings of the IEEE/CVF international conference on computer vision},
  pages={7323--7334},
  year={2023}
}

@article{Tewel2023KeyLockedRO,
    title   = {Key-Locked Rank One Editing for Text-to-Image Personalization},
    author  = {Yoad Tewel and Rinon Gal and Gal Chechik and Yuval Atzmon},
    journal = {ACM SIGGRAPH 2023 Conference Proceedings},
    year    = {2023},
    url     = {https://api.semanticscholar.org/CorpusID:258436985}
}

@inproceedings{radford2021learning,
  title={Learning transferable visual models from natural language supervision},
  author={Radford, Alec and Kim, Jong Wook and Hallacy, Chris and Ramesh, Aditya and Goh, Gabriel and Agarwal, Sandhini and Sastry, Girish and Askell, Amanda and Mishkin, Pamela and Clark, Jack and others},
  booktitle={International conference on machine learning},
  pages={8748--8763},
  year={2021},
  organization={PmLR}
}

@article{li2024mulan,
  title={Mulan: Multimodal-llm agent for progressive and interactive multi-object diffusion},
  author={Li, Sen and Wang, Ruochen and Hsieh, Cho-Jui and Cheng, Minhao and Zhou, Tianyi},
  journal={arXiv preprint arXiv:2402.12741},
  year={2024}
}

@inproceedings{li2023divide,
  title={Divide \& bind your attention for improved generative semantic nursing},
  author={Li, Yumeng and Keuper, Margret and Zhang, Dan and Khoreva, Anna},
  booktitle={34th British Machine Vision Conference 2023, {BMVC} 2023},
  year={2023}
}

@inproceedings{wang2024compositional,
  title={Compositional text-to-image synthesis with attention map control of diffusion models},
  author={Wang, Ruichen and Chen, Zekang and Chen, Chen and Ma, Jian and Lu, Haonan and Lin, Xiaodong},
  booktitle={Proceedings of the AAAI Conference on Artificial Intelligence},
  volume={38},
  number={6},
  pages={5544--5552},
  year={2024}
}

@inproceedings{ge2023expressive,
  title={Expressive text-to-image generation with rich text},
  author={Ge, Songwei and Park, Taesung and Zhu, Jun-Yan and Huang, Jia-Bin},
  booktitle={Proceedings of the IEEE/CVF international conference on computer vision},
  pages={7545--7556},
  year={2023}
}

@InProceedings{Xie_2023_ICCV,
    author    = {Xie, Jinheng and Li, Yuexiang and Huang, Yawen and Liu, Haozhe and Zhang, Wentian and Zheng, Yefeng and Shou, Mike Zheng},
    title     = {BoxDiff: Text-to-Image Synthesis with Training-Free Box-Constrained Diffusion},
    booktitle = {Proceedings of the IEEE/CVF International Conference on Computer Vision (ICCV)},
    year      = {2023},
    pages     = {7452-7461}
}

@article{rassin2024linguistic,
  title={Linguistic binding in diffusion models: Enhancing attribute correspondence through attention map alignment},
  author={Rassin, Royi and Hirsch, Eran and Glickman, Daniel and Ravfogel, Shauli and Goldberg, Yoav and Chechik, Gal},
  journal={Advances in Neural Information Processing Systems},
  volume={36},
  year={2024}
}

@inproceedings{agarwal2023star,
  title={A-star: Test-time attention segregation and retention for text-to-image synthesis},
  author={Agarwal, Aishwarya and Karanam, Srikrishna and Joseph, KJ and Saxena, Apoorv and Goswami, Koustava and Srinivasan, Balaji Vasan},
  booktitle={Proceedings of the IEEE/CVF International Conference on Computer Vision},
  pages={2283--2293},
  year={2023}
}

@inproceedings{kwon2024concept,
  title={Concept weaver: Enabling multi-concept fusion in text-to-image models},
  author={Kwon, Gihyun and Jenni, Simon and Li, Dingzeyu and Lee, Joon-Young and Ye, Jong Chul and Heilbron, Fabian Caba},
  booktitle={Proceedings of the IEEE/CVF Conference on Computer Vision and Pattern Recognition},
  pages={8880--8889},
  year={2024}
}

@inproceedings{yang2024mastering,
  title={Mastering Text-to-Image Diffusion: Recaptioning, Planning, and Generating with Multimodal LLMs.},
  author={Yang, Ling and Yu, Zhaochen and Meng, Chenlin and Xu, Minkai and Ermon, Stefano and Cui, Bin},
  booktitle={Icml},
  volume={3},
  number={6},
  pages={7},
  year={2024}
}

@article{jiang2024comat,
  title={CoMat: Aligning Text-to-Image Diffusion Model with Image-to-Text Concept Matching},
  author={Jiang, Dongzhi and Song, Guanglu and Wu, Xiaoshi and Zhang, Renrui and Shen, Dazhong and Zong, Zhuofan and Liu, Yu and Li, Hongsheng},
  journal={arXiv preprint arXiv:2404.03653},
  year={2024}
}

@article{hu2024ella,
  title={Ella: Equip diffusion models with llm for enhanced semantic alignment},
  author={Hu, Xiwei and Wang, Rui and Fang, Yixiao and Fu, Bin and Cheng, Pei and Yu, Gang},
  journal={arXiv preprint arXiv:2403.05135},
  year={2024}
}

@inproceedings{tang2023daam,
  title={What the daam: Interpreting stable diffusion using cross attention},
  author={Tang, Raphael and Liu, Linqing and Pandey, Akshat and Jiang, Zhiying and Yang, Gefei and Kumar, Karun and Stenetorp, Pontus and Lin, Jimmy and T{\"u}re, Ferhan},
  booktitle={Proceedings of the 61st Annual Meeting of the Association for Computational Linguistics (Volume 1: Long Papers)},
  pages={5644--5659},
  year={2023}
}

@inproceedings{jin2025latexblend,
        title={LatexBlend: Scaling Multi-concept Customized Generation with Latent Textual Blending},
        author={Jin, Jian and Zhenbo, Yu and Yang, Shen and Fu, Zhenyong and Yang, Jian},
        booktitle={Proceedings of the IEEE/CVF conference on computer vision and pattern recognition},
        year={2025}
}

@inproceedings{
  wang2025msdiffusion,
  title={{MS}-Diffusion: Multi-subject Zero-shot Image Personalization with Layout Guidance},
  author={Xierui Wang and Siming Fu and Qihan Huang and Wanggui He and Hao Jiang},
  booktitle={The Thirteenth International Conference on Learning Representations},
  year={2025},
  url={https://openreview.net/forum?id=PJqP0wyQek}
}

\clearpage
\appendix
\section{Dataset Construction and Evaluation Details}

\begin{figure*}[!t]
    \centering
    \includegraphics[width=0.8\linewidth]{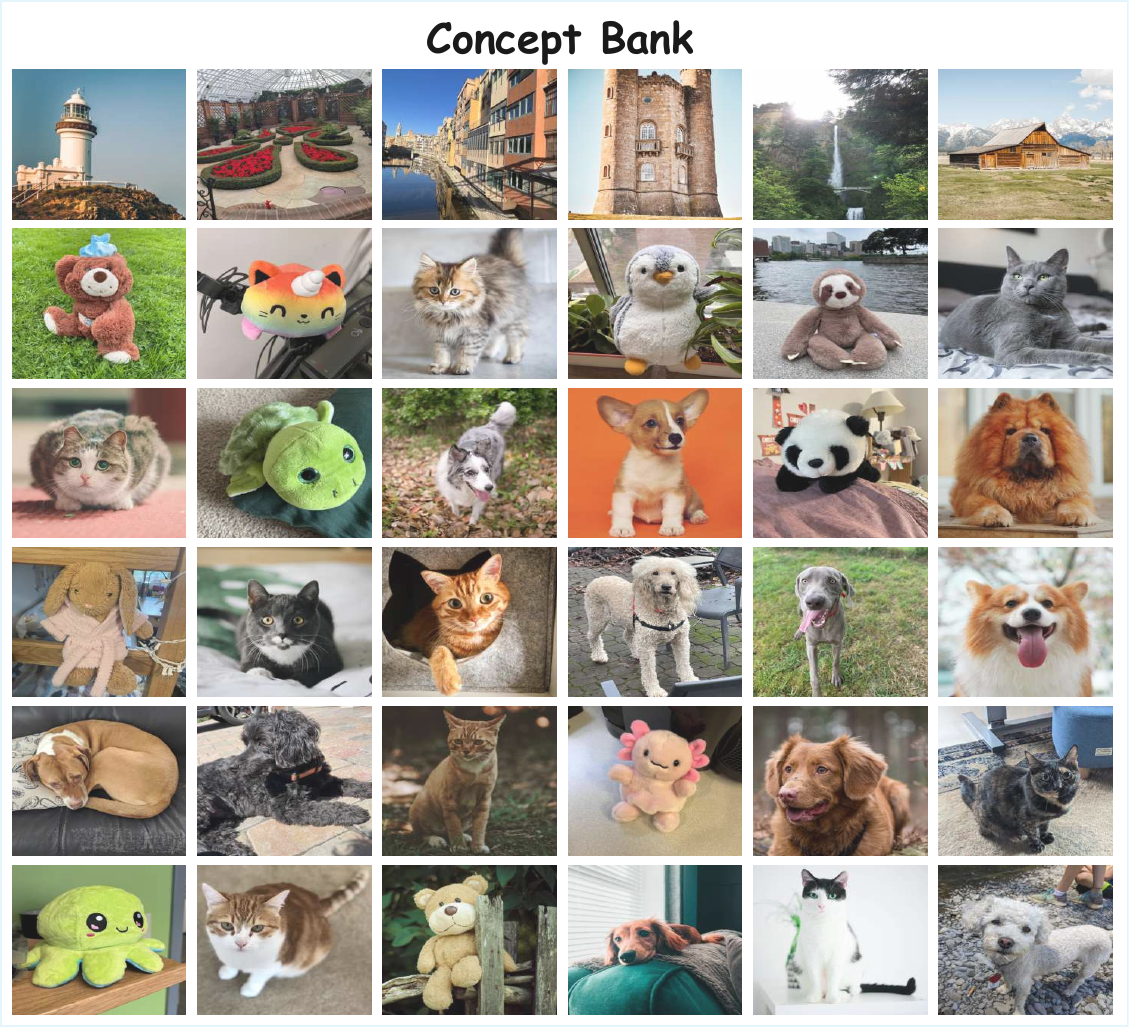}
    \caption{Overview of the 36 reference concepts used in our evaluation dataset, sourced primarily from the Concept101 and DreamBooth datasets.}
    \label{fig:dataset_overview}
\end{figure*}
\subsection{Dataset Composition and Prompt Design}
The evaluation dataset in this study is built upon a concept bank containing 36 reference images. These include 30 subject concepts (10 distinct plush toys, 9 distinct cats, and 11 distinct dogs) primarily sourced from the widely used \textbf{Concept101} and \textbf{DreamBooth} datasets, alongside 6 background concepts.

To comprehensively evaluate the model's capabilities in attribute binding and concept fusion during multi-subject generation, these subjects are combined with a predefined set of 21 distinct attributes: \textbf{goggles, scarf, bow, sunglasses, flower crown, crown, glasses, jacket, bandanas, necklace, ribbon, hat, coat, cape, bell collar, vest, helmet, headphones, tie, shorts,} and \textbf{frisbee}. We designed four distinct syntactic prompt templates for the generation:

\begin{itemize}
    \item \texttt{photo of \{theme1\} \{wearing/with a\} \{attr1\} and \{theme2\} \{action\_clause\} \{wearing/with a\} \{attr2\}, \{bg\} background}
    \item \texttt{\{theme1\} \{wearing/with a\} \{attr1\} and \{theme2\} \{wearing/with a\} \{attr2\}, both \{action\_clause\}, \{bg\} background}
    \item \texttt{photo of \{theme1\} with \{attr1\} \{action\_clause\} and \{theme2\} with \{attr2\} \{action\_clause\}, plain \{bg\} background}
    \item \texttt{a \{theme1\} \{wearing/with a\} \{attr1\} \{action\_clause\}} \\
          \texttt{next to a \{theme2\} \{wearing/with a\} \{attr2\}} \\
          \texttt{\{action\_clause\} on \{bg\} background}
\end{itemize}

During the generation process, the action clause (\texttt{\{action\_clause\}}) is randomly selected from 5 predefined actions: \textit{running, walking, sitting, dancing, and standing}. The background (\texttt{\{bg\}}) is randomly sampled from 6 predefined scenes.

\subsection{Combination Strategy and Dataset Scale}
To eliminate model bias towards specific conceptual combinations or spatial positions, we implemented a balanced combination and position-flipping strategy.

\textbf{Intra-class and Inter-class Combinations.} Subjects from the three groups (plush toys, cats, dogs) are paired pairwise. The 10 concepts within the plush toy group are uniformly paired with other plush toys, cats, and dogs.

\textbf{Attribute Allocation and Spatial Flipping.} The 21 pre-selected attributes form a large combination pool. To ensure distributional balance, we randomly sample 210 attribute pairs for each combination and swap the spatial positions (left/right) of the subjects to test spatial comprehension. Plush toy combinations utilize subsets of 140 or 210 attribute pairs depending on the paired class. Additionally, specialized subsets with fixed actions or backgrounds were constructed as control groups.

\textbf{Total Dataset Scale.} Based on the aforementioned permutation logic, the total number of generated evaluation images reaches 10,080 (calculated as $3 \times 10 \times 140 + 9 \times 210 + 11 \times 210 + 8 \times 210 = 10080$).

\subsection{Evaluation Protocol}
Given the scale of the generated dataset and the cost of utilizing Large Vision-Language Models (VLMs) for automated evaluation, we adopted a sampling evaluation mechanism. The core evaluation relies on Qwen3.5-Plus. We apply a uniform random sampling strategy to select 5,000 images from the full 10,080 generated images as the evaluation benchmark. This sampling scale ensures statistical significance while keeping the computational overhead practical.

\subsection{Rationale for Selecting Qwen3.5-Plus as the Evaluator}
In multi-subject personalized generation tasks, traditional feature-similarity metrics (e.g., CLIP-T or DINO) exhibit significant limitations. For instance, CLIP suffers from the ``bag-of-words'' phenomenon, struggling to distinguish complex compositional syntax (e.g., failing to differentiate ``a cat wearing a tie and a dog wearing a hat'' from ``a cat wearing a hat and a dog wearing a tie''). To accurately evaluate attribute binding (BIND) and attribute misalignment (MIS), we selected Qwen3.5-Plus as our Multi-modal LLM (MLLM) judge for the following reasons.

\textbf{Fine-Grained Compositional Reasoning.} Qwen3.5-Plus possesses strong spatial relationship reasoning and fine-grained visual-textual alignment capabilities. It correctly resolves topological and subordinate relationships between entities and attributes, enabling it to accurately judge whether specific attributes (e.g., clothing, accessories) are exclusively bound to their designated subjects.

\textbf{Contextual Analysis and Error Attribution.} When evaluating Attribute Misalignment (MIS), it is important to differentiate among attribute leakage, confusion, and absence. Qwen3.5-Plus can perform structured reasoning based on prompt queries (e.g., ``Does an attribute from Subject A appear in the spatial region of Subject B?''), enabling objective MIS penalty scoring that goes beyond traditional feature matching.

\textbf{Balance of Performance and Cost.} Qwen3.5-Plus demonstrates high API throughput and cost-effectiveness when processing thousands of evaluation images. Its accuracy in complex Visual Question Answering (VQA) tasks is comparable to leading closed-source models, while maintaining practical costs for large-scale validation.

\section{Analysis of Text Encoder Embedding Space and Cross-Attention Behavior}
\label{sec:semantic_deep_dive}

The main paper demonstrates the quantitative advantage of MultiCompose on MSP-Bench. This section provides additional analysis of the text encoder embedding space and the U-Net cross-attention maps to examine how our training and inference designs contribute to the observed improvements. Specifically, we use t-SNE projections to visualize the relative positions of personalized token embeddings within the text encoder's feature space, and cross-attention heatmaps to inspect spatial activation patterns during denoising.

\subsection{t-SNE Visualization of Token Embeddings}

\begin{figure}[htbp]
    \centering
    \includegraphics[width=\linewidth]{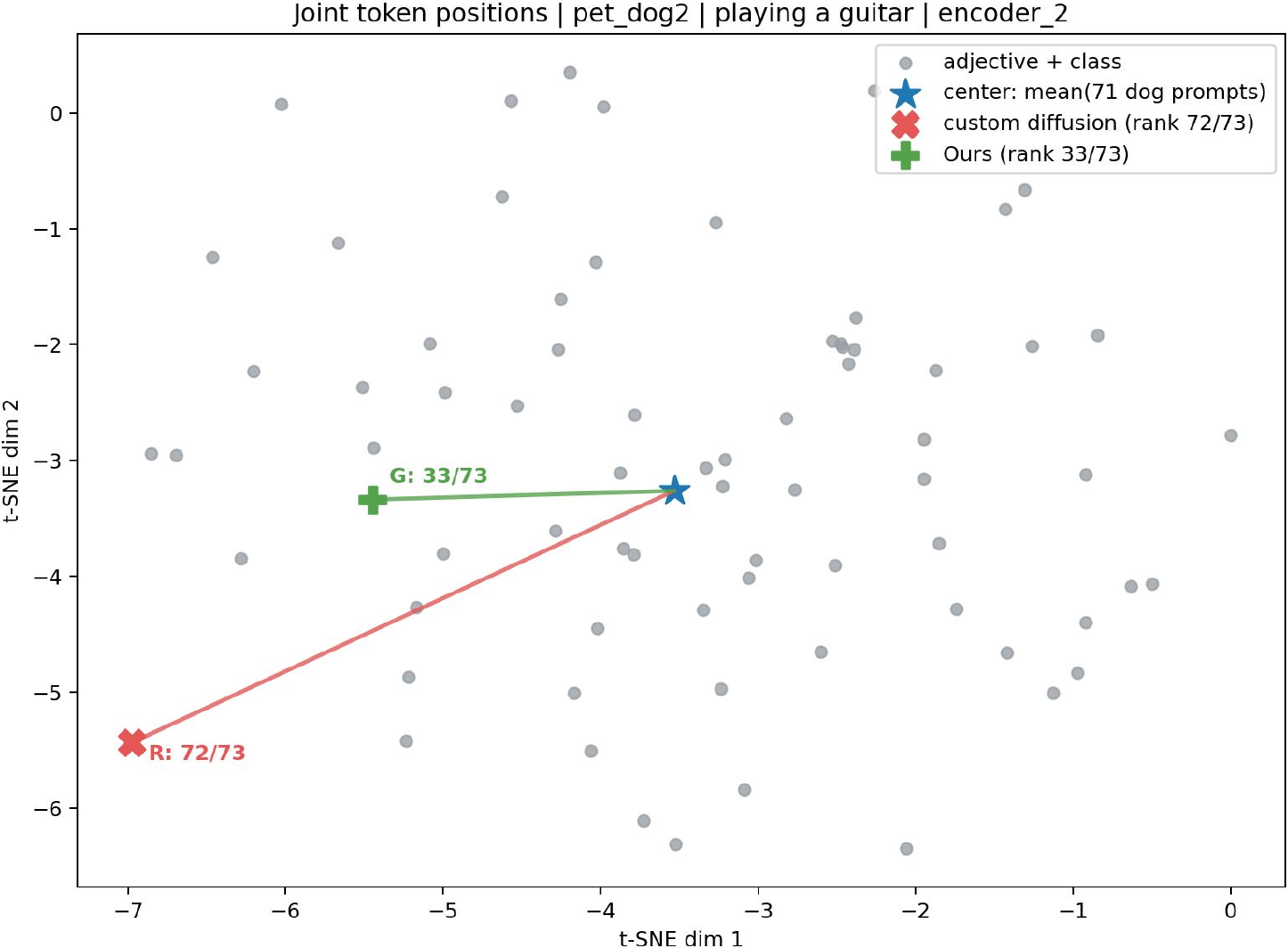}
    \caption{\textbf{t-SNE projection of token embeddings in the SDXL text encoder space.} Gray dots represent 71 natural prompts describing ``a dog playing a guitar''; the blue star marks their centroid. Custom Diffusion's token (red cross) drifts to rank 72/73, while ours (green cross) remains near the centroid at rank 33/73.}
    \label{fig:tsne}
\end{figure}

A key challenge in personalized multi-concept composition is that fine-tuning can shift a personalized token's embedding away from the category-level distribution in the text encoder's feature space. When this shift is large, the U-Net's cross-attention mechanism can no longer correctly associate the personalized token with other tokens in the prompt (e.g., attributes or actions), resulting in degraded compositionality. In Figure~\ref{fig:tsne}, we project token embeddings extracted from SDXL's \texttt{encoder\_2} into 2D via t-SNE (perplexity=30, random state=42) to visualize the embedding positions for the personalized subject \texttt{pet\_dog2} under the compositional prompt ``playing a guitar''.

\begin{figure*}[t!]
    \centering
    \includegraphics[width=\linewidth]{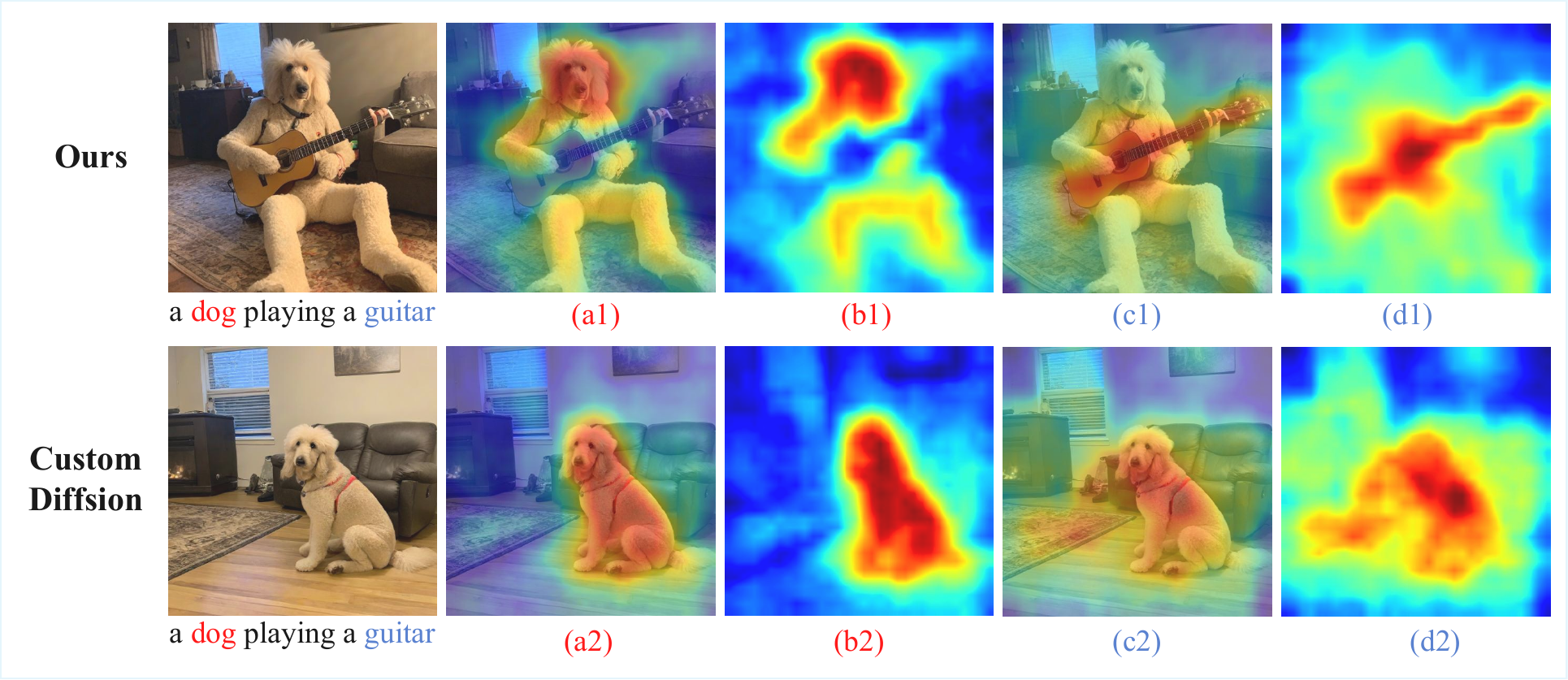}
    \caption{\textbf{Cross-attention heatmap visualization.} \textbf{Top row (a1--d1), Ours:} (a1) \texttt{<dog>} attention overlaid on the generated image; (b1) \texttt{<dog>} attention map; (c1) \texttt{<guitar>} attention overlaid; (d1) \texttt{<guitar>} attention map. Both tokens produce spatially concentrated activations confined to their respective regions. \textbf{Bottom row (a2--d2), Custom Diffusion:} The same layout, where \texttt{<dog>} attention disperses across the full image and \texttt{<guitar>} attention fails to localize the object.}
    \label{fig:heatmap}
\end{figure*}

\subsubsection{Reference Space Construction}
To establish a reference distribution, we formulated 71 natural language prompts with the structure \textit{[attribute] dog playing a guitar}, covering 70 attributes across four categories plus one attribute-free baseline: \textbf{Appearance} (small, big, cute, fluffy, brown, black, white, golden, tiny, giant, etc.), \textbf{Personality} (playful, friendly, loyal, smart, brave, gentle, calm, energetic, curious, etc.), \textbf{Status} (young, old, strong, fast, quiet, sleepy, alert, clean, healthy, wet, dry, etc.), and \textbf{Style} (beautiful, happy, elegant, graceful, charming, adorable, active, lazy, etc.).

The mean embedding of these 71 prompts defines the semantic centroid (blue star in Figure~\ref{fig:tsne}). We compute the Euclidean distance from each token embedding to this centroid and rank all 73 tokens (71 natural prompts + 2 personalized tokens), where Rank 1 indicates the closest to the centroid.

\subsubsection{Baseline: Embedding Drift in Custom Diffusion}
As shown in Figure~\ref{fig:tsne}, Custom Diffusion's personalized token (red cross) ranks 72nd out of 73, positioned far from the natural prompt cluster. This occurs because standard fine-tuning optimizes only for reconstruction loss without constraints on the embedding position. The resulting token embedding overfits to the visual appearance of the reference images and loses its association with the broader category semantics. As a consequence, the U-Net cannot properly interpret this token together with attribute tokens (e.g., ``guitar''), leading to failures in both identity preservation and attribute binding during multi-subject generation.

\subsubsection{Our Method: Balanced Embedding Position}
In contrast, our MultiCompose token (green cross) is positioned within the natural prompt cluster, achieving a rank of 33/73. The semantic preservation loss $\mathcal{L}_{sp}$ constrains the token embedding to remain close to the category-level prior during optimization. This ranking reflects a balance between two failure modes: Rank 1 would indicate collapse to the generic ``dog'' prior with loss of personalized identity, while Rank 72 (as observed in Custom Diffusion) indicates isolation from the category distribution with loss of compositionality. At Rank 33, the token retains enough distance from the centroid to encode the specific visual identity of \texttt{pet\_dog2}, while remaining close enough to preserve the compositional relationships needed for multi-attribute generation.

\subsection{Cross-Attention Heatmap Analysis}
\begin{figure*}[htbp]
    \centering
    \includegraphics[width=0.95\textwidth]{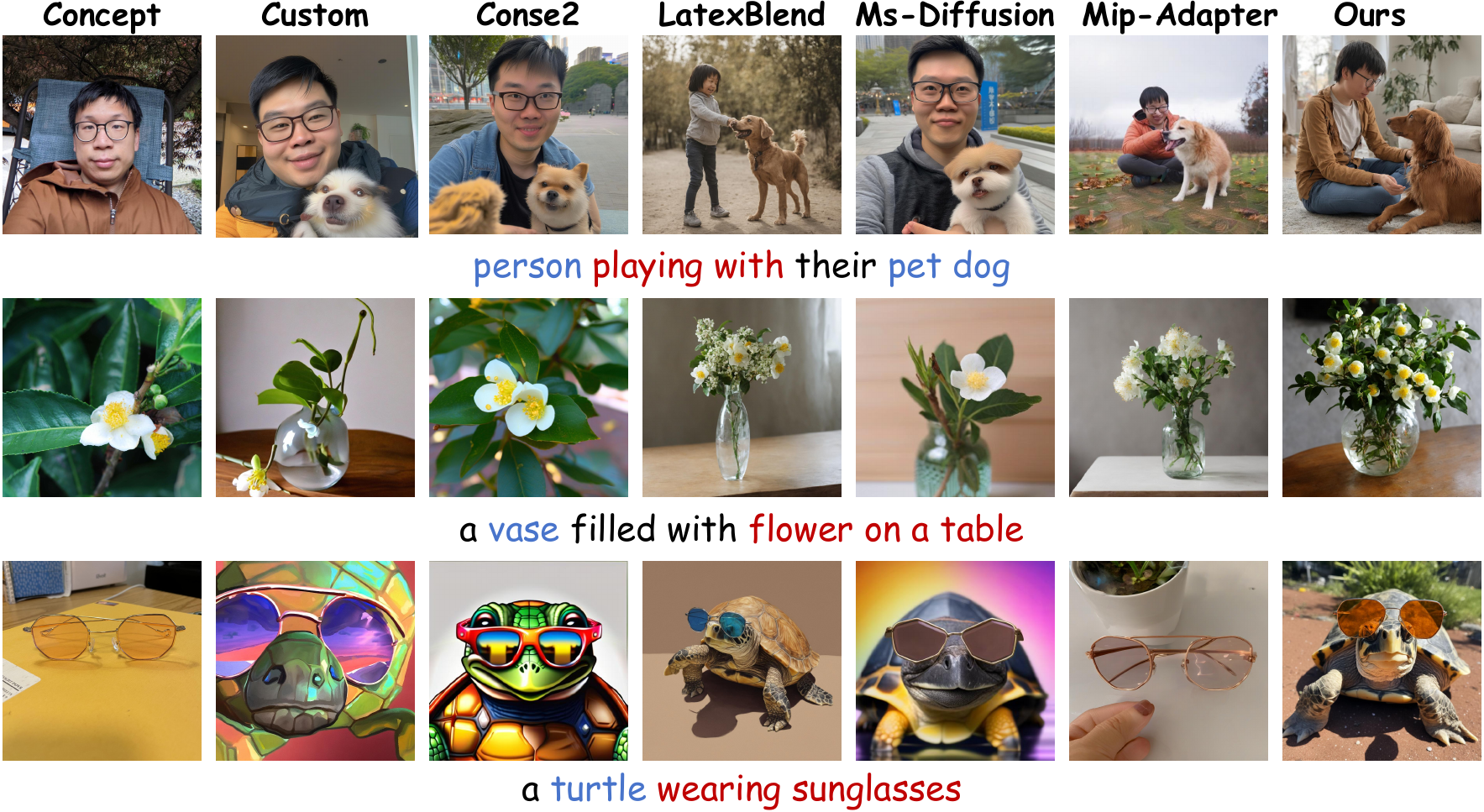}
    \caption{\textbf{Qualitative comparison on complex single-concept composition.} \textbf{Row 1 (Person):} Identity preservation during cross-category interaction. \textbf{Row 2 (Flower):} Object consistency under spatial layout constraints. \textbf{Row 3 (Turtle):} Fine-grained local attribute binding.}
    \label{fig:complex_comparison}
\end{figure*}
The embedding position in the text encoder space directly affects the cross-attention patterns in the U-Net during denoising. Figure~\ref{fig:heatmap} shows the cross-attention heatmaps for the subject token \texttt{<dog>} and the object token \texttt{<guitar>}, extracted from the middle blocks of SDXL's U-Net, averaged across layers and heads, and normalized to $[0,1]$.

As shown in Figure~\ref{fig:heatmap} (top row), our method produces well-separated cross-attention maps: the \texttt{<dog>} token concentrates on the subject region (a1, b1), and the \texttt{<guitar>} token activates only in the instrument area (c1, d1). The two activation regions exhibit minimal spatial overlap, indicating that each token attends to its semantically corresponding area without interfering with the other. This spatial precision results from two complementary mechanisms: $\mathcal{L}_{sp}$ regularization keeps the token embedding within the compositional distribution of the text encoder space, and the pre-fusion attention isolation enforces non-overlapping spatial allocation during inference.

In contrast, Custom Diffusion (bottom row) exhibits diffuse attention patterns. The \texttt{<dog>} token's activation spreads over the entire image without focusing on the subject (a2, b2), and the \texttt{<guitar>} token fails to form a coherent spatial response (c2, d2). This lack of spatial selectivity indicates that the drifted embedding cannot provide sufficient semantic grounding for the cross-attention mechanism. The resulting attention maps are consistent with the embedding drift observed in the t-SNE analysis, and correlate with the lower generation quality of Custom Diffusion in multi-subject scenarios.

\subsection{Summary}

The analysis above connects three levels of the generation pipeline. At the training stage, $\mathcal{L}_{sp}$ regularization maintains the personalized token within the category-level distribution of the text encoder embedding space. At the inference stage, the well-positioned embedding allows the attention isolation mechanism to produce spatially separated cross-attention maps. Together, these two components contribute to the improved multi-subject attribute binding observed in the main paper's quantitative evaluation.

\subsection{Visual Validation: Single-Concept Robustness}
\label{sec:visual_validation}

To further validate the generalization of our approach, we present qualitative comparisons on single-concept synthesis under complex compositional prompts.

\subsubsection{Complex Syntactic Contexts}
In Figure~\ref{fig:complex_comparison}, we compare MultiCompose with Custom Diffusion, Cones~2, LatentBlend, MS-Diffusion, and MIP-Adapter across three categories of compositional prompts.

\textbf{Cross-Category Interaction (Row 1, Person).} The prompt ``person playing with their pet dog'' requires maintaining human identity while generating a generic dog from the pretrained prior. Baselines tend to ignore the dog or merge identities. MultiCompose correctly generates both entities with distinct identities.

\textbf{Spatial Hierarchies (Row 2, Flower).} The prompt ``a vase filled with flower on a table'' requires the model to respect the spatial containment hierarchy (flower $\rightarrow$ vase $\rightarrow$ table). Standard fine-tuning often disrupts these spatial priors. Our $\mathcal{L}_{sp}$ regularization preserves the pretrained model's spatial comprehension, allowing the personalized flower to correctly follow the prepositional structure.

\textbf{Local Attribute Binding (Row 3, Turtle).} The prompt ``a turtle wearing sunglasses'' tests whether attribute tokens are correctly assigned to the target subject. Without attention isolation, attribute tokens can be misallocated to background regions or other subjects, resulting in missing or misplaced accessories. MultiCompose confines the ``sunglasses'' attribute to the turtle's head region, preventing leakage to unrelated areas of the image.

\subsubsection{24-Attribute Stress Test}
\begin{figure*}[!htbp]
    \centering
    \includegraphics[width=0.95\textwidth]{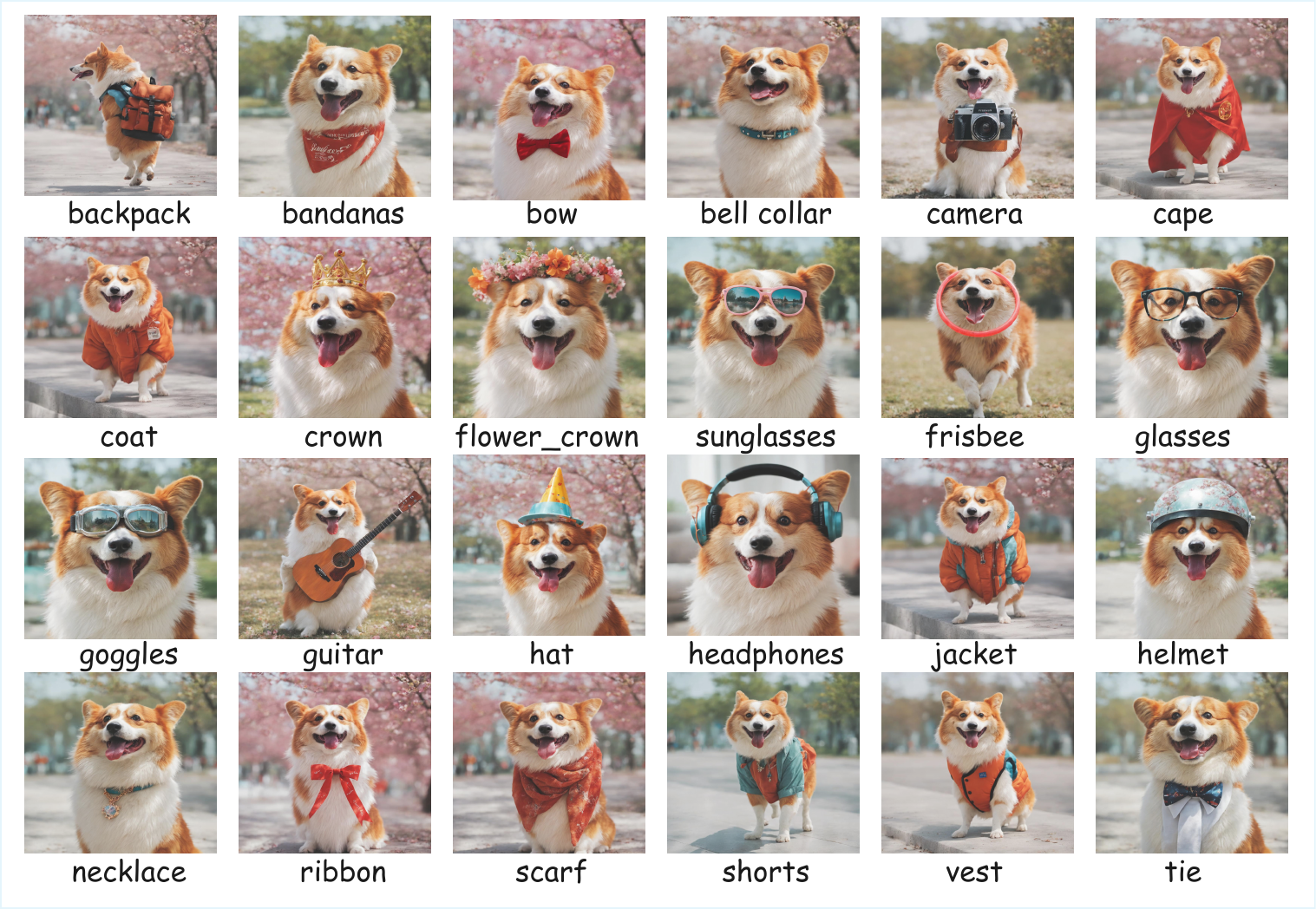}
    \caption{\textbf{24-attribute stress test.} A single personalized subject (Corgi) composed with 24 attributes across \textit{Apparel}, \textit{Wearables}, \textit{Hand-held Items}, and \textit{Decorations}. The model achieves consistent binding across all variations without corrupting the subject identity.}
    \label{fig:corgi_24}
\end{figure*}

To test the vocabulary coverage of our personalized tokens, we compose a single subject (Corgi) with 24 distinct attributes, as shown in Figure~\ref{fig:corgi_24}.

This test validates two aspects. First, for attributes that alter the subject silhouette (e.g., \textit{coat}, \textit{vest}, \textit{jacket}), the model correctly deforms the body shape while preserving the personalized fur pattern on visible regions (head, paws). Second, for attributes that require novel poses absent from the training data (e.g., \textit{camera}, \textit{guitar}), the personalized token inherits the pose priors from the base SDXL model, demonstrating that $\mathcal{L}_{sp}$ regularization preserves the compositionality of the token with the pretrained vocabulary.

\section{Multi-Subject Composition}
\label{sec:multi_subject_composition}

This section presents qualitative results on multi-subject scenarios, covering baseline comparisons, attribute disentanglement, positional control, and orthogonal factor decoupling.

\subsection{Baseline Comparisons on Heterogeneous Subject Pairs}

\begin{figure*}[htbp]
    \centering
    \includegraphics[width=\textwidth]{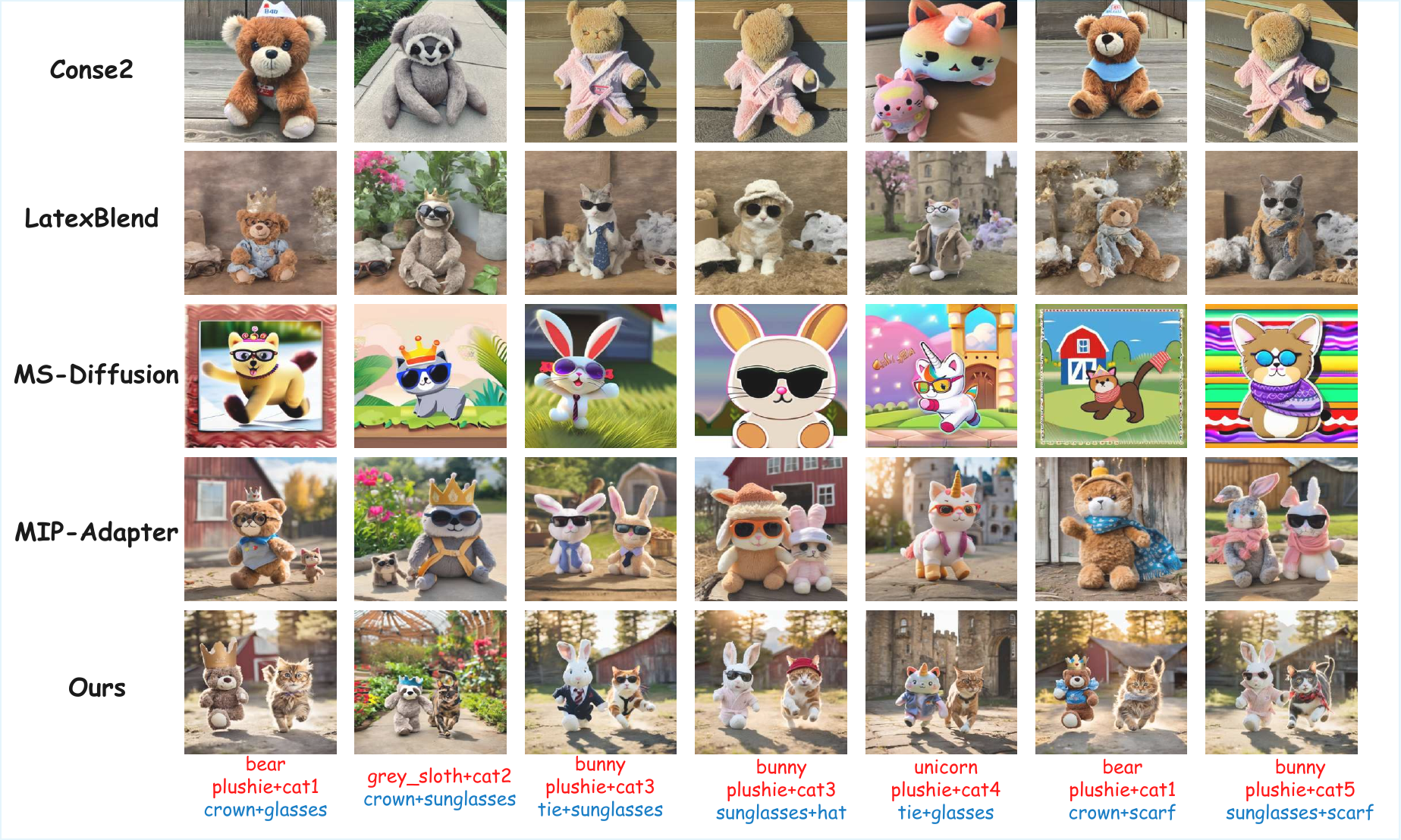}
    \caption{\textbf{Qualitative comparison: Plushie + Cat.} MultiCompose preserves the distinct textures of the plushie and the biological features of the cat while maintaining disjoint attribute binding.}
    \label{fig:plushie_cat}
\end{figure*}

\begin{figure*}[htbp]
    \centering
    \includegraphics[width=0.88\textwidth]{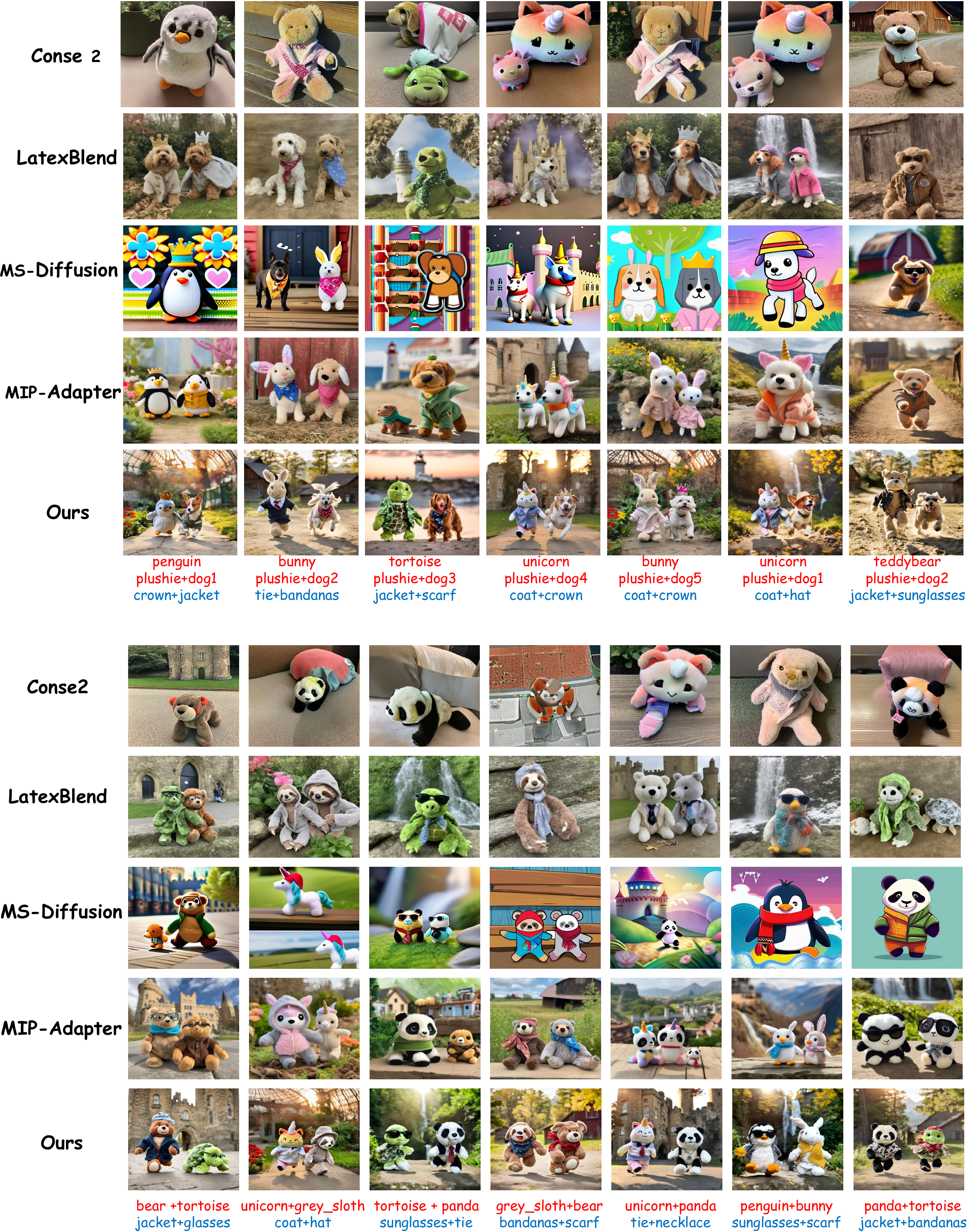}
    \caption{%
      \textbf{Qualitative comparison on heterogeneous and symmetric multi-subject pairs.}
      Each row shows one subject pair configuration; columns correspond to
      \textit{Cones~2}, \textit{LatentBlend}, \textit{MS-Diffusion}, \textit{MIP-Adapter},
      and MultiCompose (ours), from left to right.
      \textbf{(a) Plushie + Dog}: heterogeneous pairs spanning a cross-category visual domain gap.
      \textbf{(b) Plushie + Plushie}: symmetric pairs sharing the same semantic category.
    }
    \label{fig:plushie_dog_plushie}
\end{figure*}

We compare MultiCompose with \textit{Cones~2}, \textit{LatentBlend}, \textit{MS-Diffusion}, and \textit{MIP-Adapter} on three subject pair categories: Plushie + Cat (Figure~\ref{fig:plushie_cat}), and Plushie + Dog and Plushie + Plushie (Figure~\ref{fig:plushie_dog_plushie}).
\textbf{Plushie + Cat.} Figure~\ref{fig:plushie_cat} presents seven subject pair configurations, each pairing a personalized plushie with a distinct cat concept under two independently assigned attributes. Cones~2 frequently renders only a single subject, failing to co-compose both concepts in the same scene. LatentBlend produces both subjects in most cases but exhibits attribute leakage, where attributes designated for one subject appear on the other. MS-Diffusion generates outputs with a pronounced stylization shift, losing photorealistic subject fidelity. MIP-Adapter preserves subject appearance more reliably but still shows cross-subject attribute confusion in several configurations. MultiCompose produces both subjects with preserved identities and correctly assigns each attribute to its designated subject across all configurations.

\textbf{Plushie + Dog.} Figure~\ref{fig:plushie_dog_plushie}(a) presents seven configurations pairing plushie concepts with personalized dog subjects under attribute combinations spanning garments (jacket, coat) and accessories (bandanas, scarf, crown, hat, sunglasses). The cross-category gap between the artificial texture of the plushie and the organic appearance of the dog makes identity separation more demanding. Cones~2 and LatentBlend either fail to render one subject or merge subject-level appearance features. MS-Diffusion produces cartoon-style outputs with significant domain deviation. MIP-Adapter achieves better subject fidelity but shows attribute misalignment in multiple cases. MultiCompose consistently renders both subjects with distinct identities and correct attribute binding, preserving the textural difference between plushie and dog.

\textbf{Plushie + Plushie.} Figure~\ref{fig:plushie_dog_plushie}(b) evaluates a symmetric scenario in which both subjects belong to the same high-level semantic category, posing a harder identity separation problem. Seven plushie pairs with distinct attribute assignments are shown. Cones~2 and LatentBlend frequently produce a single plushie with blended appearance. MS-Diffusion renders both subjects as cartoon-style characters. MIP-Adapter partially maintains separate identities but exhibits attribute leakage between the two subjects. MultiCompose correctly assigns distinct identities and attributes to each plushie, demonstrating robustness in symmetric same-category composition.

\begin{figure*}[htbp]
    \centering
    \includegraphics[width=0.88\textwidth]{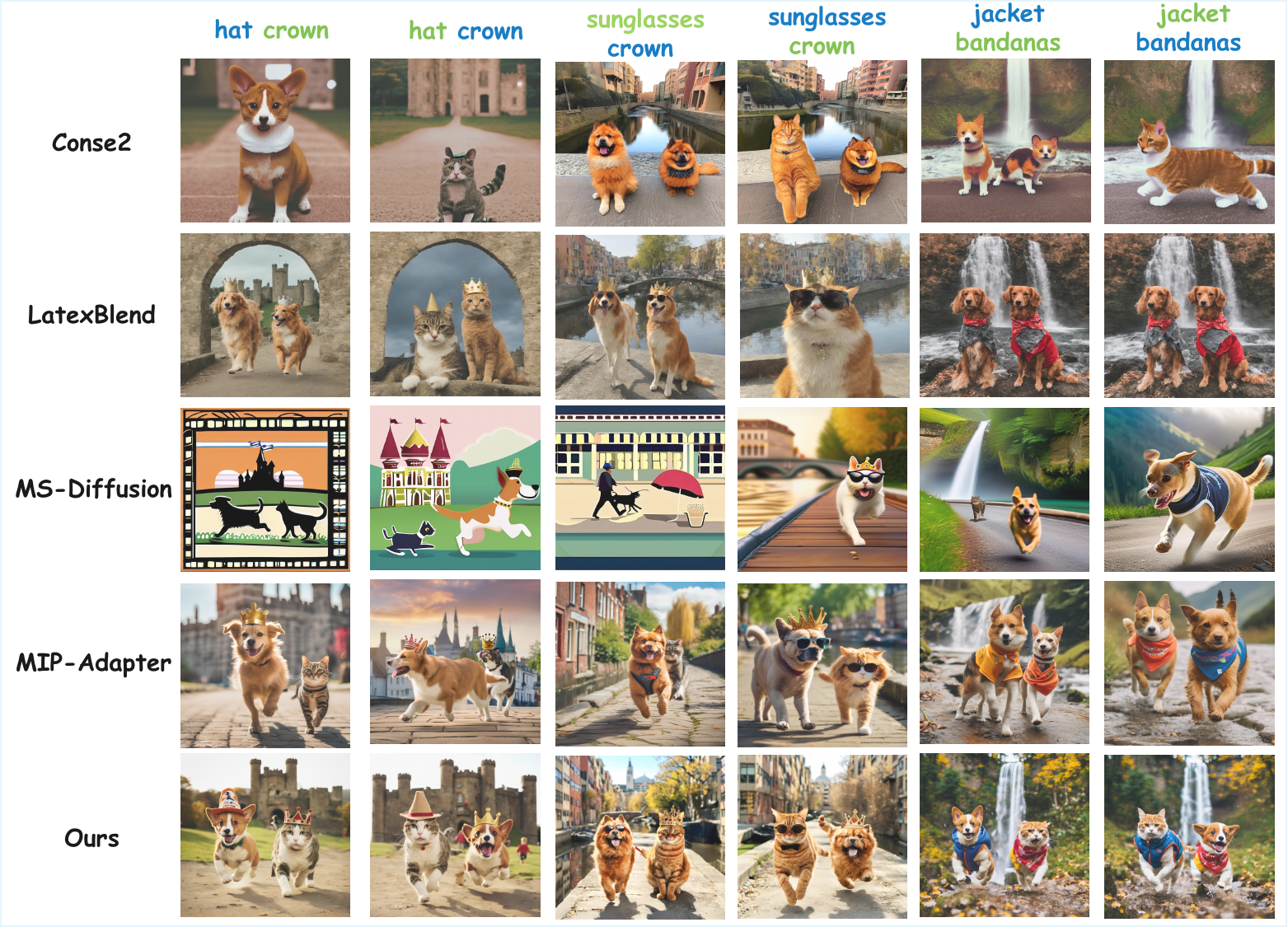}
    \caption{\textbf{Positional swapping verification.} Blue tracks the dog and its attribute; green tracks the cat and its attribute. Swapping the prompt order correctly reassigns both positions and attribute bindings.}
    \label{fig:position_swap}
\end{figure*}

\subsection{Fine-Grained Attribute Disentanglement}
A well-known challenge in multi-concept generation is the phenomenon of \textit{semantic collapse}, where linguistically or visually adjacent attributes (e.g., ``glasses'' versus ``sunglasses'', or ``bandanas'' versus ``scarf'') bleed into one another. Standard diffusion models often map these similar concepts to overlapping latent representations, causing the U-Net's cross-attention maps to activate indiscriminately across multiple subjects.

To evaluate robustness under high attribute similarity, we test the model on attribute pairs that are visually or linguistically adjacent. As shown in Figure~\ref{fig:attr_disentangle}, MultiCompose correctly assigns each attribute to its designated subject in both tested configurations. The attention isolation during the fusion phase confines each attribute token's cross-attention activation to its assigned subject region, preventing leakage to the region of a perceptually similar attribute. This confirms that the modifier tokens maintain separable positions in the text encoder embedding space under high-similarity attribute conditions.

\subsection{Spatial Layout and Positional Control}

A common failure in multi-concept generation occurs when visually similar attributes (e.g., ``glasses'' vs. ``sunglasses'', ``bandanas'' vs. ``scarf'') are confused with each other. Standard diffusion models tend to map similar attribute tokens to overlapping latent representations, causing the cross-attention maps to activate on incorrect subject regions.

\begin{figure*}[!htbp]
    \centering
    \includegraphics[width=0.85\textwidth]{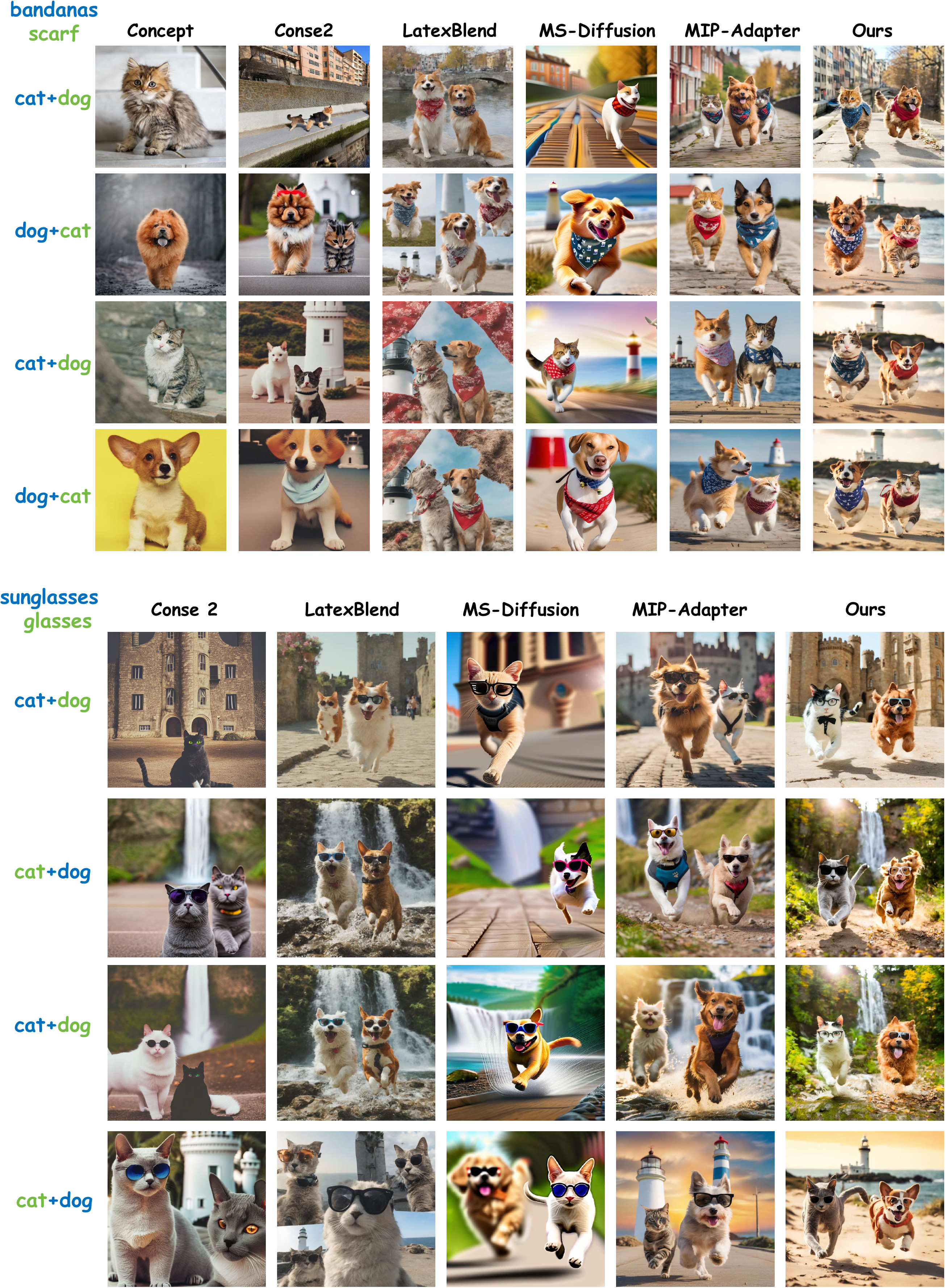}
    \caption{\textbf{Attribute disentanglement.} \textbf{Top:} Bandanas vs.\ Scarf. Blue indicates the bandana on the dog; green indicates the scarf on the cat. \textbf{Bottom:} Sunglasses vs.\ Glasses. Blue indicates sunglasses; green indicates standard glasses.}
    \label{fig:attr_disentangle}
\end{figure*}

To test the robustness of our approach against such confusion, we evaluate on attribute pairs with high visual similarity. As shown in Figure~\ref{fig:attr_disentangle}, MultiCompose correctly separates similar attributes between subjects. The attention isolation during the fusion phase prevents the texture of one attribute (e.g., ``scarf'') from affecting the region designated for another (e.g., ``bandana''), confirming that the personalized tokens maintain distinct positions in the embedding space.

Models often overfit to spatial biases in the training data, fixing certain subjects to specific positions. When the prompt syntax swaps the order of subjects, baseline models may either ignore the positional change or move the subjects without their bound attributes.

Figure~\ref{fig:position_swap} shows positional swapping results. By altering the syntactic order in the prompt, we verify that MultiCompose correctly reassigns both subject positions and their associated attributes. The pre-fusion phase interprets the spatial structure of the prompt and routes the layout accordingly, confirming that attribute binding follows the prompt syntax rather than static spatial priors.

\subsection{Orthogonal Decoupling of Generation Factors}
\label{sec:orthogonal_decoupling}

An effective compositional framework should support independent control over four factors: subject identity, attribute, action, and background. In standard diffusion models, these factors are coupled: varying the background can shift subject appearance, and changing the action can cause attribute bindings to disappear from the generated output. We conduct controlled experiments to verify that MultiCompose decouples these factors.

\subsubsection{Environmental and Subject Modularity}

\begin{figure}[!htbp]
    \centering
    \includegraphics[width=\linewidth]{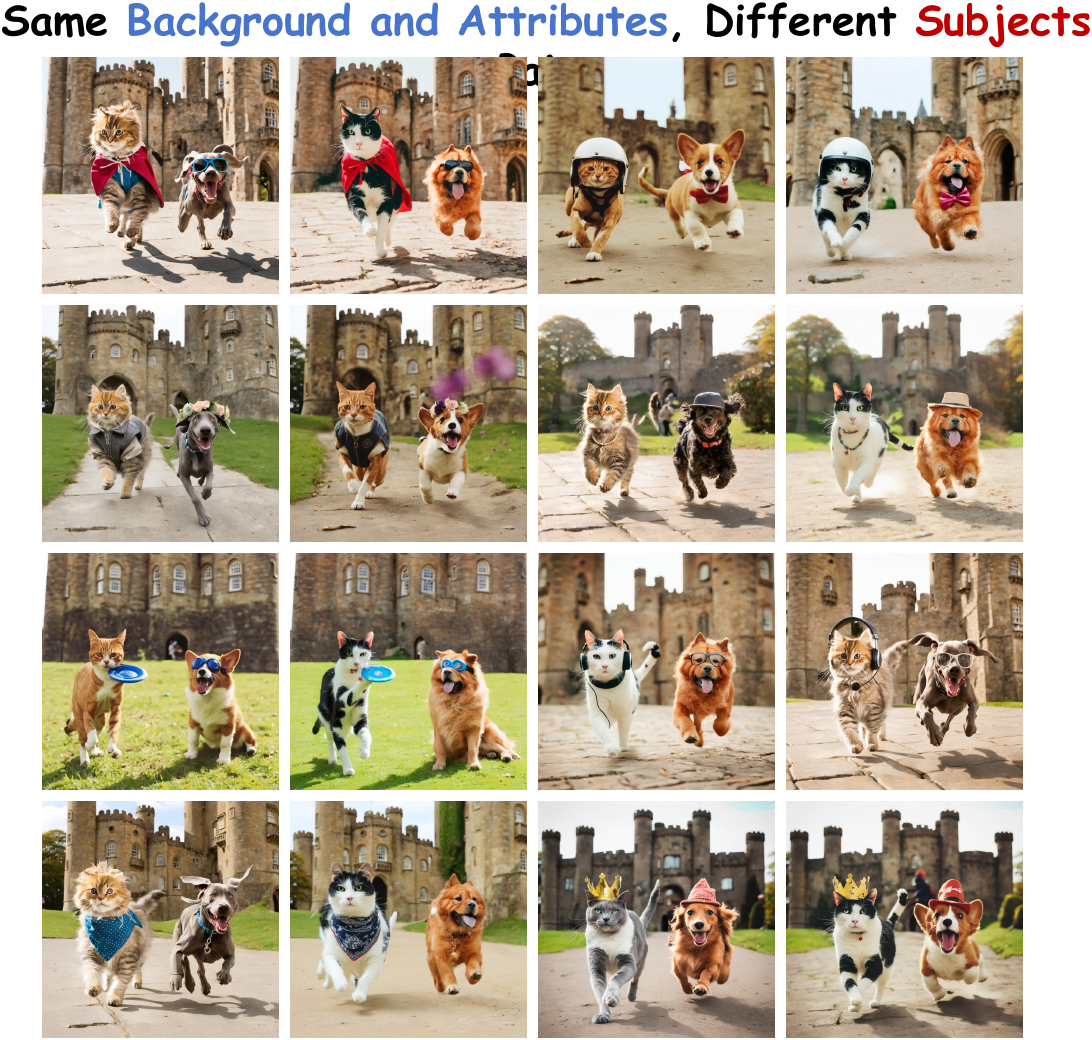}
    \caption{\textbf{Background invariance.} Same subject pair and attribute configuration across four background scenes (coastal, castle, garden, waterfall).}
    \label{fig:bg_invariance}
\end{figure}

\textbf{Background Invariance.} Figure~\ref{fig:bg_invariance} evaluates whether attribute bindings are preserved under background variation. We fix a subject pair and four attribute configurations (jacket + sunglasses, jacket + glasses, scarf + sunglasses, and jacket + glasses) while varying the background across four scenes: coastal, castle, garden, and waterfall. Subject identities and per-subject attribute assignments remain consistent across all background contexts, indicating that the fusion-phase mask routing isolates subject-level attribute control from global scene context.

\begin{figure}[!htbp]
    \centering
    \includegraphics[width=\linewidth]{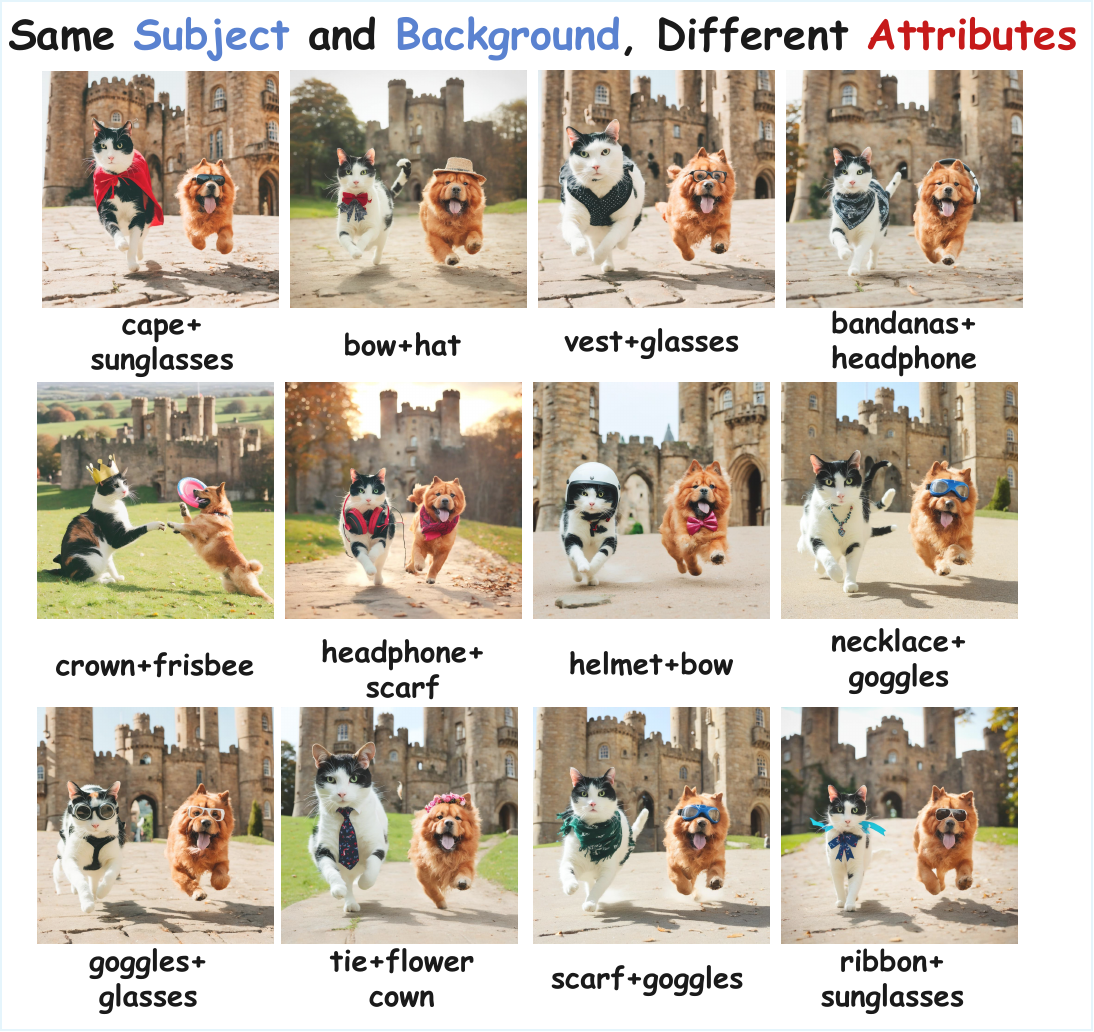}
    \caption{\textbf{Global interchangeability.} Rows: four fixed attribute pairs. Columns: four subject-background combinations.}
    \label{fig:global_interchange}
\end{figure}

\textbf{Global Interchangeability.} Figure~\ref{fig:global_interchange} tests whether attribute binding generalizes when both subjects and backgrounds are simultaneously varied. Four attribute pair configurations (jacket + sunglasses, jacket + glasses, crown + hat, and bandanas + crown) are each applied to four different subject-background combinations, covering distinct cat and dog identities across different scene contexts. The correct per-subject attribute assignment is preserved across all configurations, confirming that the attention isolation mechanism does not depend on specific subject identities or background scenes.

\begin{figure}[!htbp]
    \centering
    \includegraphics[width=\linewidth]{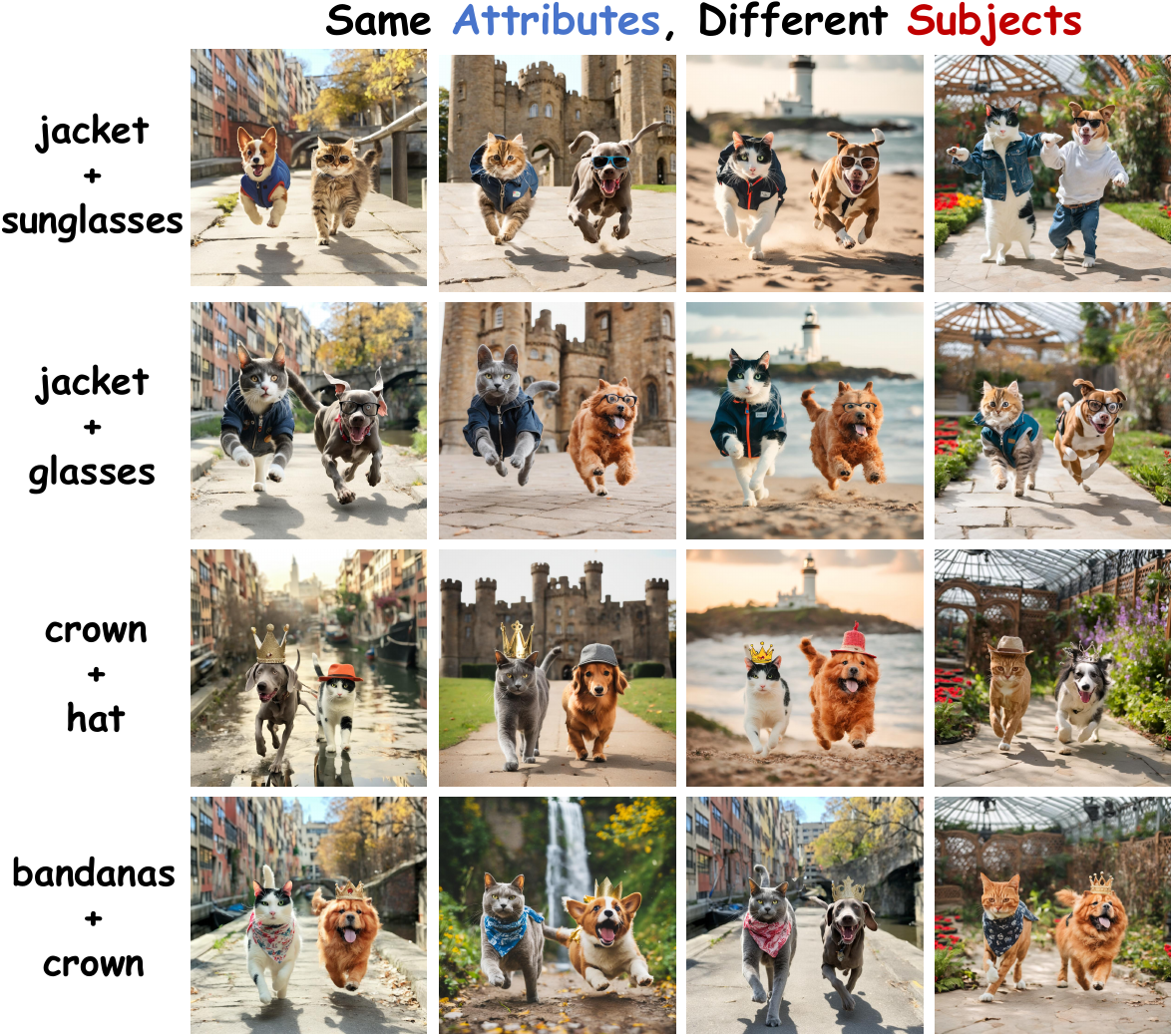}
    \caption{\textbf{Subject interchangeability.} Fixed castle background; different personalized subject pairs per column.}
    \label{fig:subject_interchange}
\end{figure}

\textbf{Subject Interchangeability.} Figure~\ref{fig:subject_interchange} fixes the background (castle) and applies multiple attribute pair configurations to different personalized cat-dog subject pairs in each column. The mask routing correctly adapts to the varying body proportions and spatial layouts of each subject pair, assigning each attribute to its designated subject without cross-subject leakage.

\begin{figure}[!htbp]
    \centering
    \includegraphics[width=\linewidth]{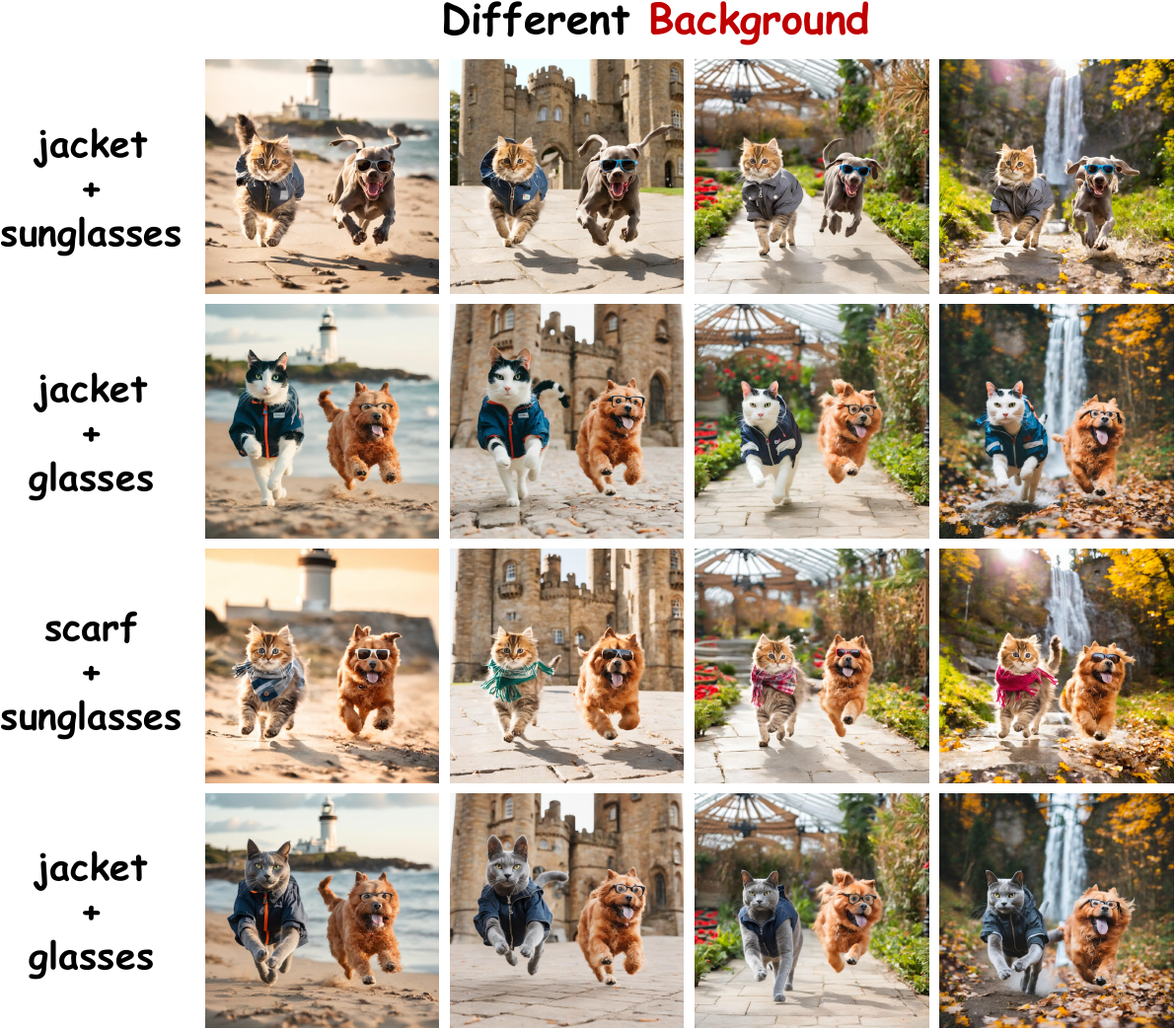}
    \caption{\textbf{Attribute versatility.} Fixed subject pair and castle background; 12 attribute pair combinations covering garments, headwear, accessories, and held objects.}
    \label{fig:attr_versatility}
\end{figure}

\begin{figure}[!htbp]
    \centering
    \includegraphics[width=\linewidth]{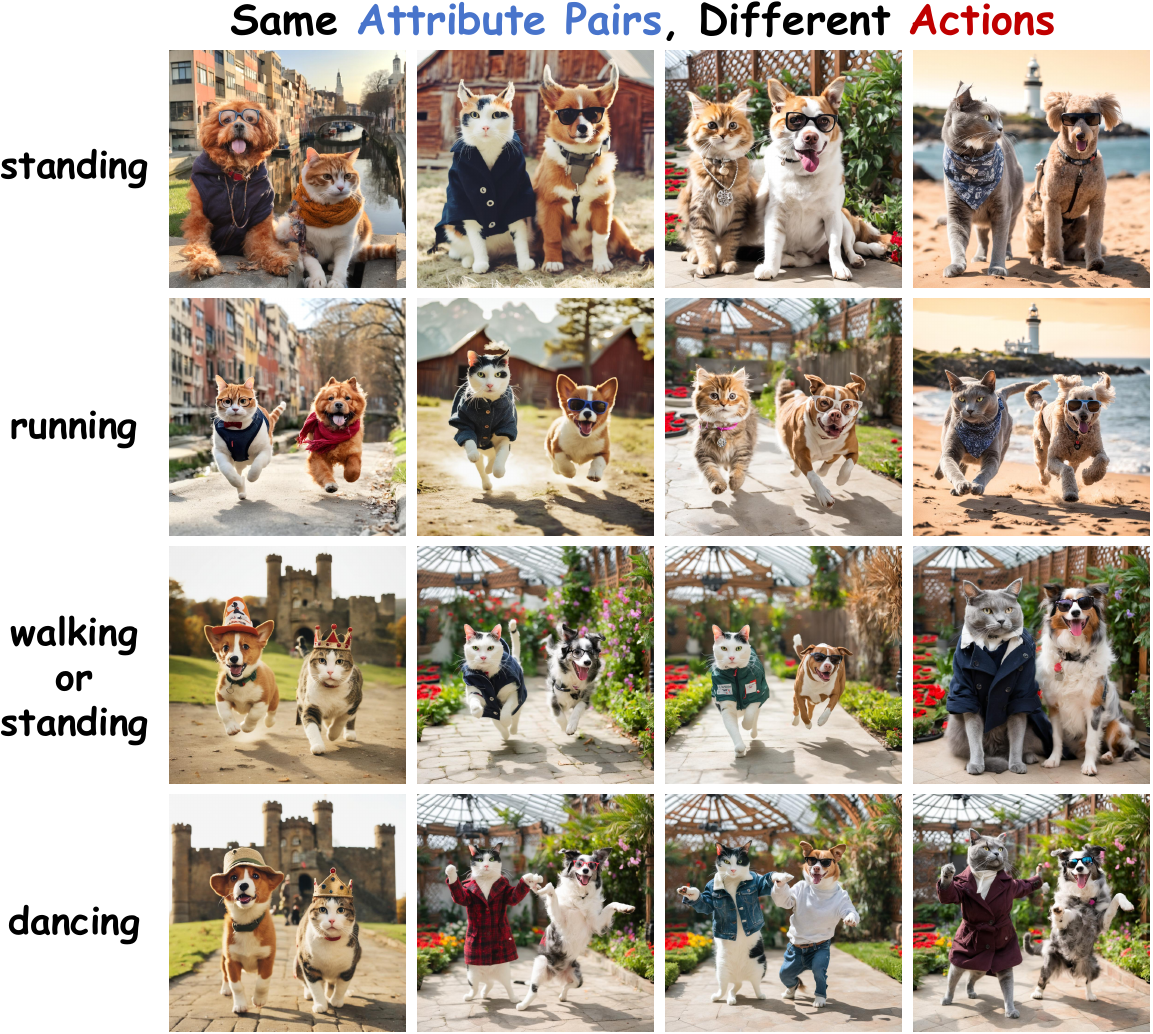}
    \caption{\textbf{Action decoupling.} Rows: four action categories (standing, running, walking, dancing). Columns: four subject-background combinations with fixed attribute pairs.}
    \label{fig:action_decouple}
\end{figure}

\subsubsection{Action Dynamics and Attribute Versatility}

\textbf{Action Decoupling.} Figure~\ref{fig:action_decouple} evaluates attribute preservation under pose variation. Four action categories are tested (standing, running, walking, and dancing) across four subject-background combinations. Dynamic actions (running, dancing) produce substantially different body poses and spatial layouts compared to static poses (standing). Attribute bindings are maintained across all four action categories, as the pre-fusion phase recomputes layout masks conditioned on the denoised layout at $t_{\mathrm{cond}}$, adapting to pose-induced spatial changes without disrupting per-subject attribute assignments.

\textbf{Attribute Versatility.} Figure~\ref{fig:attr_versatility} fixes a single personalized cat-dog subject pair and a castle background, then applies 12 distinct attribute pair combinations drawn from a 17-attribute vocabulary spanning garments (cape, vest), headwear (bow, hat, crown, helmet), accessories (sunglasses, glasses, goggles, scarf, necklace, tie, headphone, bandanas, ribbon), and held objects (frisbee). Each attribute is correctly assigned to its designated subject across all 12 configurations, demonstrating broad vocabulary coverage without subject identity degradation or cross-subject leakage.

\clearpage

\end{document}